\documentclass[11pt,fleqn]{article}
\usepackage{epsf}
\usepackage{amsmath}
\usepackage{amssymb}
\usepackage{color}
\usepackage{graphicx}
\usepackage{cite}
\usepackage{float}
\usepackage{color} 
\definecolor{darkblue}{rgb}{0,0,.5}
\usepackage{setspace} 
\usepackage{rotating}
\usepackage{xspace}

\usepackage[labelfont=bf,labelsep=period,justification=raggedright]{caption}
\usepackage{url}

\makeatletter
\renewcommand{\@biblabel}[1]{\quad#1.}
\makeatother

\date{}

\usepackage{algorithmic}
\usepackage{algorithm}
\renewcommand{\algorithmiccomment}[1]{/* #1 */}

\newcommand{\argmax}{\operatornamewithlimits{argmax}}
\newcommand{\ea}{\emph{et al.}\xspace}
\newcommand{\eg}{\emph{e.g.}\xspace}
\newcommand{\ie}{\emph{i.e.}\xspace}

\newcommand{\Lee}{\emph{Lee}\xspace}
\newcommand{\Chuang}{\emph{Chuang}\xspace}
\newcommand{\Taylor}{\emph{Taylor}\xspace}

\usepackage{lscape}
\usepackage{amssymb,epsf,amsmath}
\begin{document}

% Title must be 150 characters or less
\begin{flushleft}
{\Large
% \textbf{A comparison of network and pathway based feature extraction methods for the outcome prediction of breast cancer} 

% \vspace{0.5cm}\textbf{A comparative study of network and pathway-based classifiers versus a single gene classifier for the prediction of breast cancer outcome} 

\vspace{0.5cm}\textbf{A critical evaluation of network and pathway based classifiers for outcome prediction in breast cancer}}
% Insert Author names, affiliations and corresponding author email.

\begin{center}
%- Version \today \\
 
\end{center}

\bf{%%
C Staiger$^{1, 2, *}$, 
S Cadot$^2$,
R Kooter$^{3}$, 
M Dittrich$^{4}$,
T M\"uller$^{4}$,
GW Klau$^{1, 5, +, *}$,
LFA Wessels$^{2, 3, 6, +, *}$
}

\vspace*{0.5cm}
$^1$ Centrum Wiskunde \& Informatica, Life Sciences Group, Science Park 123, 1098 XG Amsterdam, Netherlands
\\
$^2$ Bioinformatics and Statistics, The Netherlands Cancer Institute, Plesmanlaan 121, 1066 CX Amsterdam, Netherlands
\\
$^3$ Delft Bioinformatics Lab, Faculty of Electrical Engineering, Mathematics and Computer Science, 2600 GA Delft, Netherlands
\\
$^4$ Department of Bioinformatics, Biocenter, Am Hubland, 97074 University of W\"urzburg, Germany
\\
$^5$ Netherlands Institute for Systems Biology, Amsterdam, Netherlands
\\
$^6$ Cancer Systems Biology Center, The Netherlands Cancer Institute, Plesmanlaan 121, 1066 CX Amsterdam, Netherlands
\\
$^\ast$ E-mail: c.staiger@cwi.nl, gunnar.klau@cwi.nl, l.wessels@nki.nl
\\
$^+$ shared last authorship
\end{flushleft}

% Please keep the abstract between 250 and 300 words
\section*{Abstract}
% Please keep the Author Summary between 150 and 200 words
% Use first person. PLoS ONE authors please skip this step. 
% Author Summary not valid for PLoS ONE submissions.   
%
% \todo{whole document: 
%  \begin{itemize}
%   \item terminology:
%    \begin{itemize}
%     \item pathways, gene sets, networks: secondary data
%     \item features sets derived from secondary data and gene expression: composite features 
%     \item classifiers: composite feature classifiers 
%    \end{itemize}
%   \item figure captions
%   \item Nomenclature for feature selection + secondary source: Chuang-netC
% \item WHENEVER we say ``not different'' or ``different'' add a sentence about corresponding statistical test (+ significance level). Currently have for only selection of these statements.
%  \end{itemize}
% }
%
Recently, several classifiers that combine primary tumor data, like gene expression data, and secondary data sources, such as protein-protein 
interaction networks, have been proposed for predicting outcome in breast cancer. In these approaches, new composite features are typically constructed by aggregating the  expression levels of several genes. The secondary data sources are employed to guide this aggregation. Although many studies claim 
that these approaches improve classification performance over single gene classifiers, the gain in performance is difficult to assess. 
This stems mainly from the fact that different breast cancer data sets and validation procedures are employed to assess the performance. 
Here we address these issues by employing a large cohort of six breast cancer data sets as benchmark set and 
by performing an unbiased evaluation of the classification accuracies of the different approaches. Contrary to previous claims, we 
find that composite feature classifiers do not outperform simple single gene classifiers. We investigate the effect of (1) the number of selected 
features; (2) the specific gene set from which features are selected; (3) the size of the training set and (4) the heterogeneity of the data set 
on the performance of composite feature and single gene classifiers. Strikingly, we find that randomization of secondary data sources, which 
destroys all biological information in these sources, does not result in a deterioration in performance of composite feature classifiers. 
Finally, we show that when a proper correction for gene set size is performed, the stability of single gene sets is similar to the stability 
of composite feature sets. Based on these results there is currently no reason to prefer prognostic classifiers based on composite features over 
single gene classifiers for predicting outcome in breast cancer. 
Supplementary data can be downloaded from \url{http://homepages.cwi.nl/~staiger/supplement.pdf} .

\section*{Introduction}
%%%%%%%%%%%%%%%%%%%%%%%%%%%%%%%
%%        background         %%
%%%%%%%%%%%%%%%%%%%%%%%%%%%%%%%

Modern high-throughput methods provide the means to observe genome wide changes in gene expression patterns in breast cancer samples. 
Gene expression signatures have been proposed \cite{Veer2002, Wang2005} to predict prognosis in breast cancer patients, but were shown to vary 
substantially between data sets. One possible explanation for this effect is that the data sets on which the predictors are trained are 
typically poorly dimensioned, consisting of many more genes than samples. Integrating secondary data sources like, for example, protein-protein 
interaction (PPI) networks, co-expression networks or pathways from databases such as KEGG, has recently been proposed to overcome variability 
of prognostic signatures and to increase their prognostic performance \cite{Chuang2007, Lee2008, Taylor2009, Ma2010, Abraham2010}. 
Many of these studies claim that combining gene expression data with secondary data sources to construct composite features results in higher accuracy in outcome prediction 
and higher stability of the obtained signatures. 
In addition, inclusion of the secondary sources raises the hope that the obtained signatures will be more interpretable and thus provide more 
insight into the molecular mechanisms governing survival in breast cancer.

The underlying idea of these methods is that genes do not act in
isolation, and that complex diseases such as cancer are actually
caused by the deregulation of complete processes or pathways,
representing `hallmarks of cancer' \cite{Hanahan2011}. This is
unlikely to happen due to an aberration in a single gene, and often
multiple genes need to be perturbed to disable a process. This leads
to the notion that aggregating gene expression of functionally linked
genes smooths out noise and provides more power to detect
deregulation of complete functional units and hence to obtain a clearer picture of the biological process underlying
tumorigenesis and disease outcome.

The observed improvement in classification accuracy achieved by the approaches employing secondary data is hard to assess since it is dependent 
on many factors such as the specific data sets and evaluation protocol employed. To shed more light on this issue we performed an extensive 
comparison of a simple, single gene based classifier with  three of the most popular approaches that include secondary data sources in the 
construction of the classifier. More specifically, we included the approaches proposed by Chuang \ea \cite{Chuang2007}, 
Lee \ea \cite{Lee2008} and Taylor \ea \cite{Taylor2009}. We investigated how these methods perform 
with respect to classification accuracy and stability of the set of features included in the classifiers. We will now briefly outline how 
the approaches work and point out some of the claims made by the authors. Detailed descriptions are provided in the Methods section. 

Chuang \ea \cite{Chuang2007} describe a greedy search algorithm on PPI networks. For each defined subnetwork, a composite feature is 
defined as a variant of the average of the expression values of the genes included in the subnetwork.
The score that guides the search is the 
association of the composite feature with patient outcome. Significance testing and a feature selection step are employed to select the set of 
composite features employed in the final classifier. The authors claim that classification based on subnetwork markers improves prediction 
performance on two breast cancer data sets. Moreover, they state that subnetwork markers are more reproducible across different breast cancer 
studies than single gene markers.

%Lee \ea \cite{Lee2008} employ the pathways as defined in KEGG \cite{Kanehisa2010} and the gene sets from databases like MsigDB 
Lee \ea \cite{Lee2008} employ gene sets from the MsigDB database
\cite{Subramanian2005} as secondary data source. The association of the composite feature with patient outcome is used as performance criterion, 
and a greedy search is employed to select a subset of genes from a gene set to constitute the composite feature. The value of the composite feature is derived from the expression values of the subset of genes as defined in Chuang \ea 
\cite{Chuang2007}. In contrast to 
Chuang \ea, Lee \ea do not exploit the connectivity of the pathway in the construction of the composite features. 
Lee \ea claim that by using these pathway activities a higher classification performance can be achieved on different 
cancer types, most notably leukemia, lung and breast cancer. 
They also report a higher overlap between genes in the top scoring composite features compared to the top scoring single genes.

The underlying assumption in the study by Taylor \ea \cite{Taylor2009} is that disease-causing perturbations disturb the organization 
of the interactome, which then has an effect on outcome. They
concentrate on highly connected proteins, so-called hubs, as these proteins act as 
organizers in the molecular network. In contrast to Lee \ea and Chuang \ea, Taylor \ea detect aberrations in the 
correlation structure between a hub and its immediate interactors. As correlation between two genes cannot be assessed for a single sample, 
Taylor \ea employ the pairwise expression difference between the hub and each of its interactors as features for the classifier.
While no claims are made regarding performance improvements, we included this approach in the 
comparison as it is a recently proposed, novel approach for exploiting secondary data sources to predict outcome in breast cancer.

Table~\ref{tab:methods_overview} provides a summary of the characteristics of all methods included in the comparison. It lists a description 
of each approach, the secondary data sources employed, the types of (composite) features and how the value of a (composite) feature is 
computed for a single tumor.

%%During the last years many pathway and network based methods have been developed. 

All three studies listed above use their own specific cross-validation
(CV) protocol and evaluate their method on different (combinations of) data 
sets. This makes it hard to assess the improvement over other methods. In this work, we therefore employ an unbiased training and validation  
protocol and present a comprehensive evaluation of cross data set classification performance and stability on six publicly available breast 
cancer data sets.  
Given that these classifiers are intended to predict the unknown outcome of a patient, we suggest a cross-validation procedure 
that does not assume any knowledge about the samples used for
testing. Thus, we strictly separate the training data set from the
test data set, \ie composite feature construction, the selection of the optimized number of features for classification and the training of the final classifier 
are all performed on the training data set, while the testing of this trained classifier is performed on a completely separate test set without calibrating the 
classifier on the test data. See Figure~\ref{fig:cv} and Algorithm~\ref{algo:cv} for details. In other words, in contrast to previous studies, we strictly distinguish 
between training and test data.

To prevent biases associated with a specific secondary data source, we tested the algorithms on different types of secondary data sources. (See the Materials and Methods section for detailed descriptions of all these data sources.) 
We also used two different classifier types, the nearest mean classifier (NMC) and logistic regression (LOG) to
evaluate the influence of the classifier on prediction performance. We chose these classifiers since Popovici \ea \cite{Popovici2010} 
confirmed earlier findings that these classifiers performed best on various breast cancer related classification tasks. Similarly, different 
feature extraction strategies were employed. While the included set of feature extraction approaches is by no means exhaustive, we employed 
approaches that were shown to perform well on gene expression based diagnostic problems \cite{Wessels2005}. All evaluations were performed 
on a curated collection consisting of six breast cancer cohorts \cite{Reyal2008} including the cohort from the Netherlands Cancer Institute\cite{Vijver2002}.

%%% further contributions %%%%

In contrast to earlier findings we find that when we apply a proper correction for the number of genes appearing in the composite features 
employed by the composite feature classifier, the stability of single gene feature sets is comparable to the stability of composite feature 
sets. Much to our surprise, and in contrast to other studies, we also find that integrating secondary data, as done in the evaluated methods, 
does \emph{not} lead to increased classification accuracy when compared to simple single gene based methods. Our findings are partly consistent 
with the findings of Abraham \ea \cite{Abraham2010}, where the authors show that averaging over gene sets does not increase the 
prediction performance over a single gene classifier.

We investigated several possible factors that may explain the disappointing performance of approaches incorporating secondary data. First, we 
looked into the effect of the way the number of features is selected. Second, we looked into the effect of the exact size 
and composition of the starting gene set. This factor could play a role since not all genes are included in secondary data sources, hence 
classifiers employing secondary data sources may be at a disadvantage compared to single gene classifiers that select the gene set from all 
genes on the chip. Third, we investigated the effect of sample
size. Finally, we looked into the effect of heterogeneity of the data sets on 
classifier performances. We find that none of these factors change our general findings. 

In addition to all these technical factors, we also investigated whether the biological information captured in the secondary data contributes 
to the classification performance of the composite feature classifiers. 
To our astonishment we found that composite classifiers constructed from 25 \emph{randomized} versions of the secondary data sources performed 
comparably to composite classifiers trained on the original, non-randomized data. 

We conclude that further research has to be done on finding effective ways to integrate secondary data sources in predictors of outcome in 
breast cancer. In order to facilitate this research, and to ensure a standardized and objective way of establishing improvements over 
state-of-the-art approaches, we make all the code, data sets and
results employed in this comparison available for download and use
upon request.

\begin{table}[hbtp]\small
\begin{tabular}[ht]{p{1.3cm}p{1cm}p{5cm}p{1.6cm}p{1.6cm}p{3cm}} \hline
  {\bf Method} & {\bf Symbol} & {\bf Description} & {\bf Secondary}\ {\bf data} & {\bf Feature} & {\bf Feature value}\\ \hline
 Single genes & SG & \raggedright Calculates t-statistic between mRNA expression distributions of the two patient groups &
  None & Single gene & mRNA expression of the gene \\ \hline 
\raggedright Lee\ \newline \ea \cite{Lee2008} & L,\newline \Lee & \raggedright Calculates for each pathway a set of genes with high t-statistic between the averaged 
gene expression and the two patient groups & MsigDB, KEGG &
\raggedright Subset of genes in a pathway & Averaged mRNA expression
of genes in set\\ \hline
\raggedright Chuang\ \ea \cite{Chuang2007} & C, \newline \Chuang &
\raggedright Calculates subnetworks with high mutual information between the averaged gene expression 
of the genes in the subnetworks and the class labels of the two patient groups & KEGG, HPRD,
I2D, NetC & \raggedright Genes in a subnetwork & Averaged mRNA expression of the genes in the network \\ \hline
\raggedright Taylor \newline \ea \cite{Taylor2009} & T, \newline \Taylor &
\raggedright Finds hub proteins that, given the two patient groups, show different Pearson correlation 
of the mRNA expression between the hub proteins and all of their
direct interactors & KEGG, HPRD, I2D, NetC & \raggedright Edge between a hub and its interactor & \raggedright Difference of mRNA expression of hub and its interactors (edge weights) 
\end{tabular} 
\caption{Overview of evaluated feature extraction methods.}
As secondary data sources we used the KEGG database \cite{Kanehisa2010}  and the C2 data set of the MsigDB \cite{Subramanian2005}, as PPI data 
we used the information from KEGG, HPRD \cite{Prasad2009} and the OPHID/I2D databases \cite{Brown2005}. In addition we used the PPI network published by Chuang \ea \cite{Chuang2007} (NetC).
\label{tab:methods_overview}
\end{table}

\section*{Results}

\subsection*{Current composite feature classifiers do not outperform single gene classifiers on six breast cancer data sets}
\label{sec:SGbetter}
\begin{figure}[tbp]
\begin{minipage}{0.5\linewidth}
\textbf{A}\\
\includegraphics[width=3.5in]{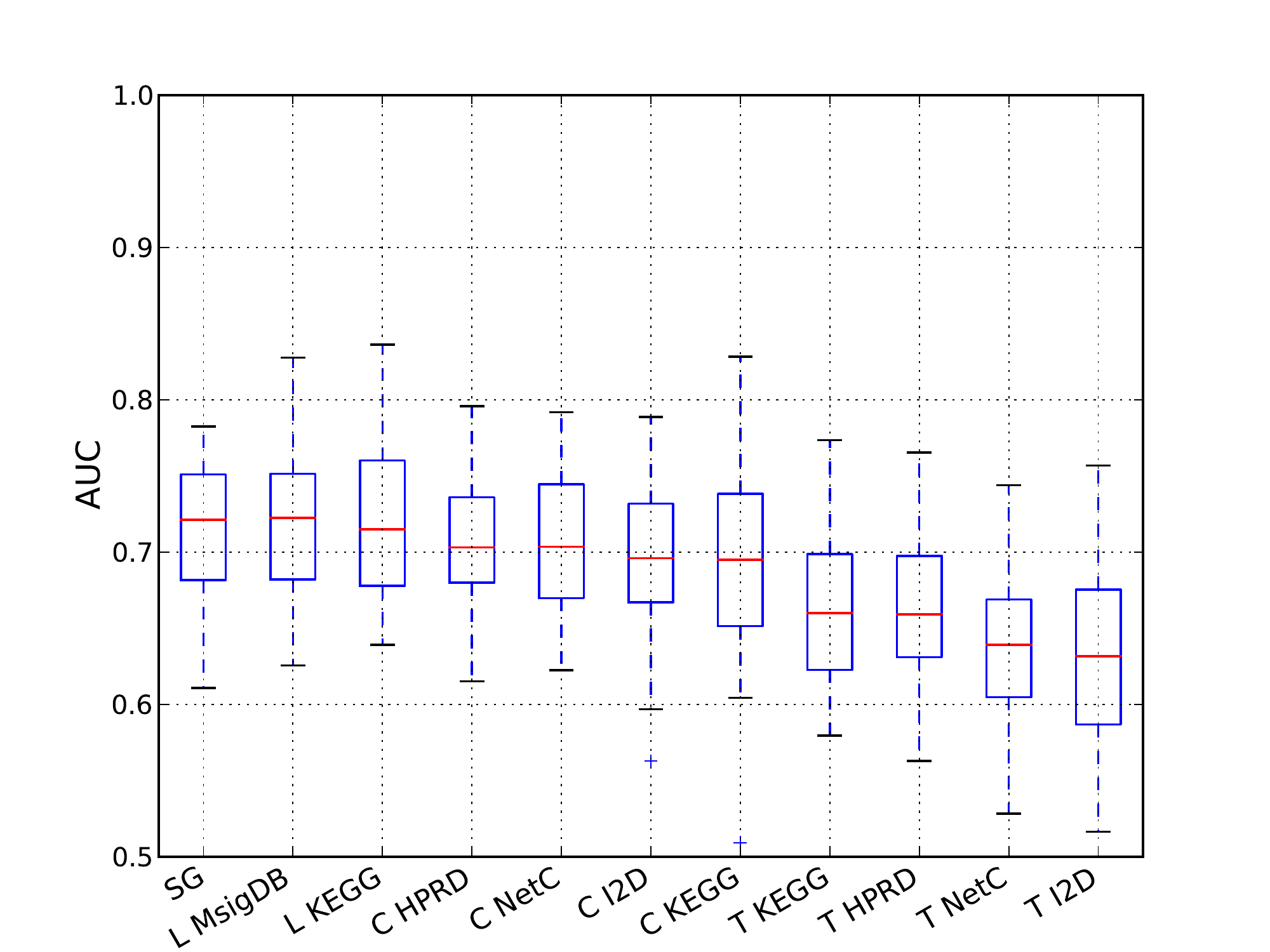} 
\end{minipage}
\begin{minipage}{0.5\linewidth}
\textbf{B}\\
\includegraphics[width=3.5in]{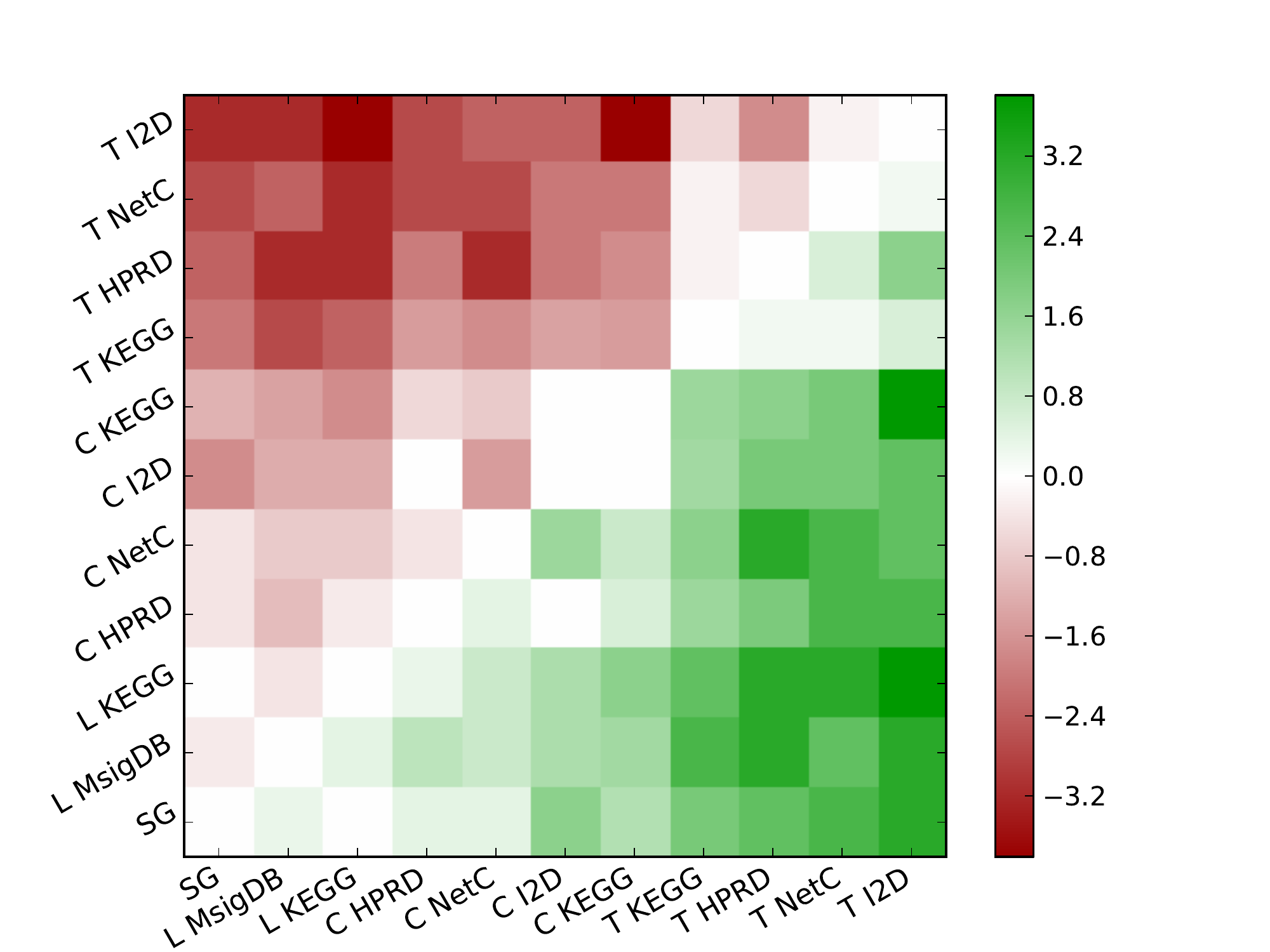}
\end{minipage}
\caption{\small
{\bf Performances of the NMCs employing single genes and composite features constructed from  different secondary data sources.} 
For each combination of feature extraction method and secondary data source and each pair of data sets we obtained one AUC value resulting in 30 AUC 
values per combination. The number of features for each classifier was determined in the cross-validation procedure (CV-optimized). 
\textbf{A:} Each box plot shows the median, the $25\%$ and $75\%$ percentiles and the standard deviation of the 30 
AUC values. 
Outliers are depicted by crosses. The boxes are sorted in descending order according to the median. \textbf{B:} This panel shows the result 
of pairwise 
comparisons between all combinations of feature extraction methods and secondary data sources. If, for a given combination of training and test data set, 
the AUC value of classifier 
$i$ is higher (lower) than the AUC value of classifier $j$ on the same
test data set, it is counted as a win (loss) for classifier $i$. Element $(i,j)$ 
in the matrix represents the $\log_2$ 
%% obacht. for final version. i
%% understand that it is now too
%% much work\todo{g: which log? or even better: figure legend as ``normal'' numbers, not log} %% g: added log_2 for nw
ratio of wins to losses of method $i$ compared to method $j$. Green indicates an overall win, red an overall 
loss and white represents draws. The rows and columns are sorted as in
Panel A. {\bf Abbreviations:} SG: Single genes;
C: \Chuang; L: \Lee and T: \Taylor. %%\todo{g: check win/loss
%%  values. really only max.\ 15 wins/losses?} g: done. is log _ratio_. 
}
\label{fig:auc_boxplot}
\end{figure}

We compared the performance of a nearest-mean classifier (NMC) using single genes with a NMC employing feature extraction methods based on 
pathway and PPI data. The results are depicted in Figure~\ref{fig:auc_boxplot}. For each combination of 
secondary data source and feature extraction approach, Figure~\ref{fig:auc_boxplot}A shows the box plots of the area under the receiver-operator characteristics curve (AUC) values obtained for each pair of data sets -- using one data set of the pair as training set and the other data set in the pair as test set. The feature extraction approaches are ranked in descending order based on the median AUC values. The box plots suggest that no composite 
classifier performs better than the single gene classifier. 
Indeed, testing whether the mean performance of the single gene classifier is different from the mean performance of any composite classifier reveals that there is no difference (null hypothesis can not be rejected) except for \Taylor and \Chuang-I2D, where the single gene classifier is clearly superior. See Table~S1 for details.
This fact is confirmed by the pairwise comparisons between all
classifiers, see Figure~\ref{fig:auc_boxplot}B.
A green square means that the combination in the row won more frequently over the combination in the column across the data set pairs. 
The good performance of the single gene classifier is reflected by the fact that the bottom row does not contain a single red box.
Also, the generally poor performance of \textit{Taylor} is clearly reflected in the 
dark red rows associated with this approach.

We also provide the classification results for the LOG classifiers in Figure~S1 and Table~S2. In general, the performances are lower than for the NMC, with the best combination not even reaching an AUC of 0.7 while several NMC classifiers clearly exceed 0.7. Apart from \Lee all 
composite LOG classifiers
perform equally or even worse than the single gene LOG
classifier. However, it should be noted that the performance of the
LOG classifier is highly variable as a function of the number of
included features---see Figures~S2-S4. In addition, the
training procedure does not converge for all feature values as is
evident in the AUC vs.\ number of features curves that end
abruptly. %\todo{l: do not like this phrase: 'Thus, it is not possible
          %to distinguish if this is a trend or merely 
           %an artifact.'} %%g: deleted it
The high sensitivity to the number of features is most evident
for the \Taylor composite NMCs. Clearly, the LOG classifier as employed here (and as employed by Chuang \ea) requires additional regularization to ensure convergence across the whole range of feature values. Also in combination with this classifier, \Taylor performed poorly. This together with the high computational burden associated with this method, prompted us to omit \Taylor from the remaining analyses.

Based on the results presented in Figure~\ref{fig:auc_boxplot}, we conclude that on the six breast cancer data sets employed in this comparison, 
composite classifiers employing secondary data sources do not outperform single gene classifiers on the task of 
predicting outcome in breast cancer, provided that a robust single gene classifier is employed.

\subsection*{Four hypotheses regarding the lack of observed performance differences}
Next we formulated a number of hypotheses that could explain why classifiers employing secondary data sources do not  
outperform single gene classifiers. These hypotheses relate to (1) the feature selection approach employed; (2) the 
starting set of genes employed in each of the approaches; (3) the effect of the training set size on performance and (4) the homogeneity of the 
data set employed. In the following sections we will investigate these hypotheses one by one.

%%%%%%%%%%%%%%%%%%%%%%%%%%%%%%%%%%%%%%%%%%%%%%%
% Hypotheses: (all related to the algorithm)
%	CV procedure (NOT INCLUDED IN THIS VERSION)
%	Type of classifier (NOT INCLUDED IN THIS VERSION)
%	Feature optimization (50,100,150)
%	Starting set
%	Pooling
%   Heterogeneity (ER)
%%%%%%%%%%%%%%%%%%%%%%%%%%%%%%%%%%%%%%%%%%%%%%%

\subsubsection*{The number of selected features does not effect relative performances}

\begin{figure}[tbp]
\centering
\includegraphics[width=5in]{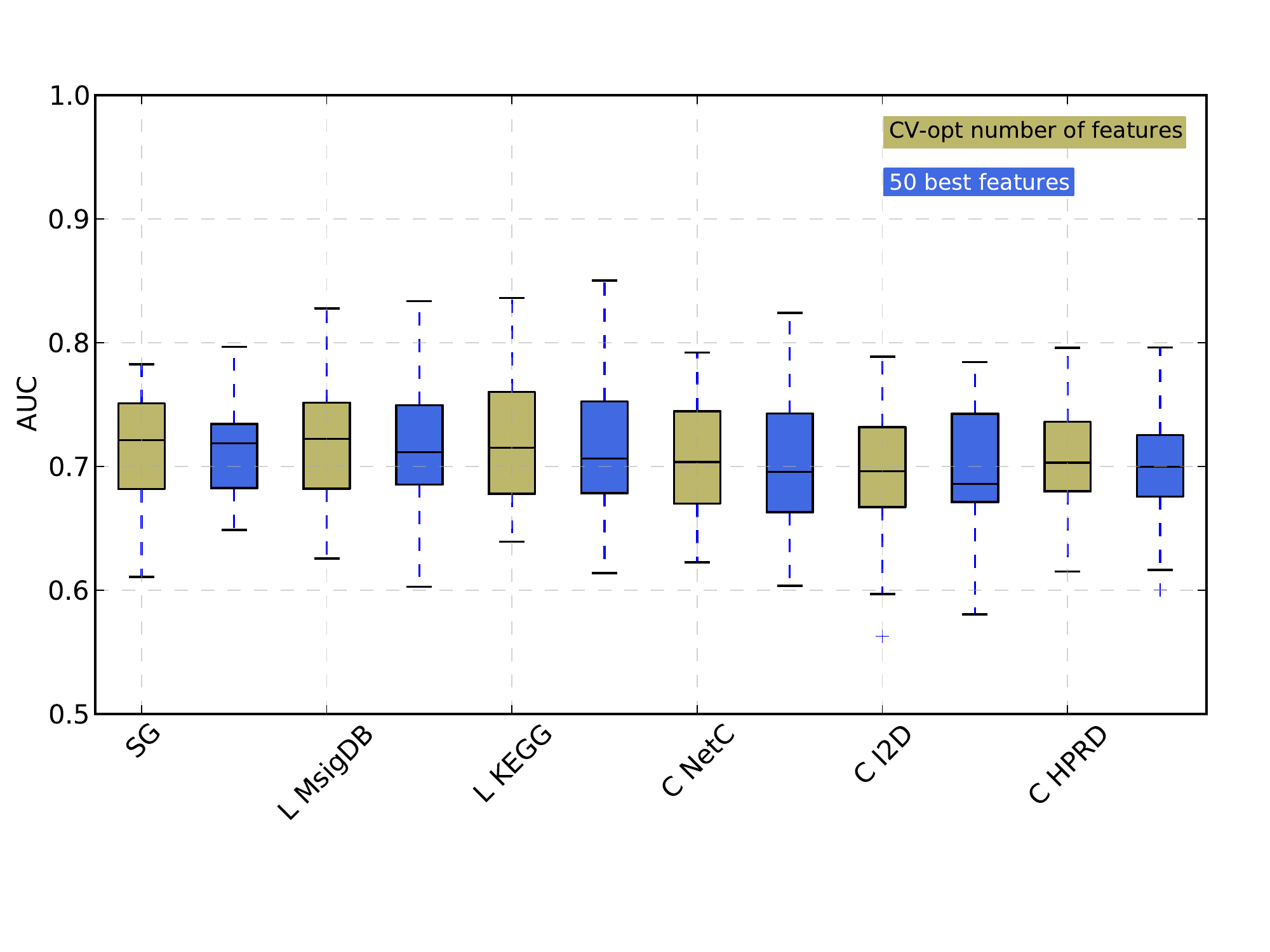} 
\caption{\small{\bf Performance of all classifiers restricted to 50 features.} Comparison of the performance of the classifiers when the 
number of features is trained in the CV procedure (denoted as `CV-opt number of features', same values as in Figure~\ref{fig:auc_boxplot}) and 
when the 50 best scoring features (denoted as `50 best features') are selected for classification. We cannot show the values 
for \Chuang-KEGG, \Taylor-KEGG or \Taylor-HPRD since for some data sets, the number of significant composite features was lower than 50.
Abbreviations of methods as in Figure~\ref{fig:auc_boxplot}.
}
\label{fig:boxplot50}
\end{figure}

In the cross-validation protocol that we proposed for unbiased performance evaluation and also employed in the comparison, we employ individual 
feature   filtering to select an optimized number of features to employ in the classifier. While this approach is sub-optimal, we (Wessels \ea \cite{Wessels2005}) and others have shown that these simple approaches perform the best in predicting phenotypes based on gene expression data. However, we observed in the curves showing the AUC values as a function of the number
of ranked features included in the classifier (Figures S2-S4) that the 
AUC values for the NMC are very stable across a large range of features for most approaches, and that the absolute maximal AUC value chosen 
during the feature selection routine might only marginally differ from the performance obtained with other feature values. For this reason, and 
since the 
selection of the optimized number of features introduces additional variability between the approaches, we decided to fix the number of 
features to 50, 100 and 150 for most approaches. We chose these values as they covered the feature ranges across which the performance remained 
stable in all approaches. The results for fixing the number of features to 50 are depicted in Figure~\ref{fig:boxplot50} while results for 100 
and 150 features are presented in Figures S5 and S6. 
%Most classifiers benefit from fixing the number of features and show higher mean performance. 
When accounting for multiple testing, no classifier using a fixed number of features performs significantly different from its counterpart using the number 
of features determined by cross-validation. See p-values of the pairwise Wilcoxon rank test in Tables S4, S6 and S8. As expected, there are only minor differences between the performance of classifiers when the number of features is
restricted to 50, 100 and 150 (Tables S3, S5 and S7) with any significant differences favoring single genes. This confirms that the number of features is not a critical parameter. Based on these results, we can 
conclude that the number of selected features does not explain the observed differences between composite feature classifiers and single gene 
classifiers.

\subsubsection*{Restricted gene sets are not detrimental to composite feature classifiers}

\begin{figure}[tb]
\centering
\begin{minipage}{0.3\linewidth}
A\\
\includegraphics[width=\linewidth]{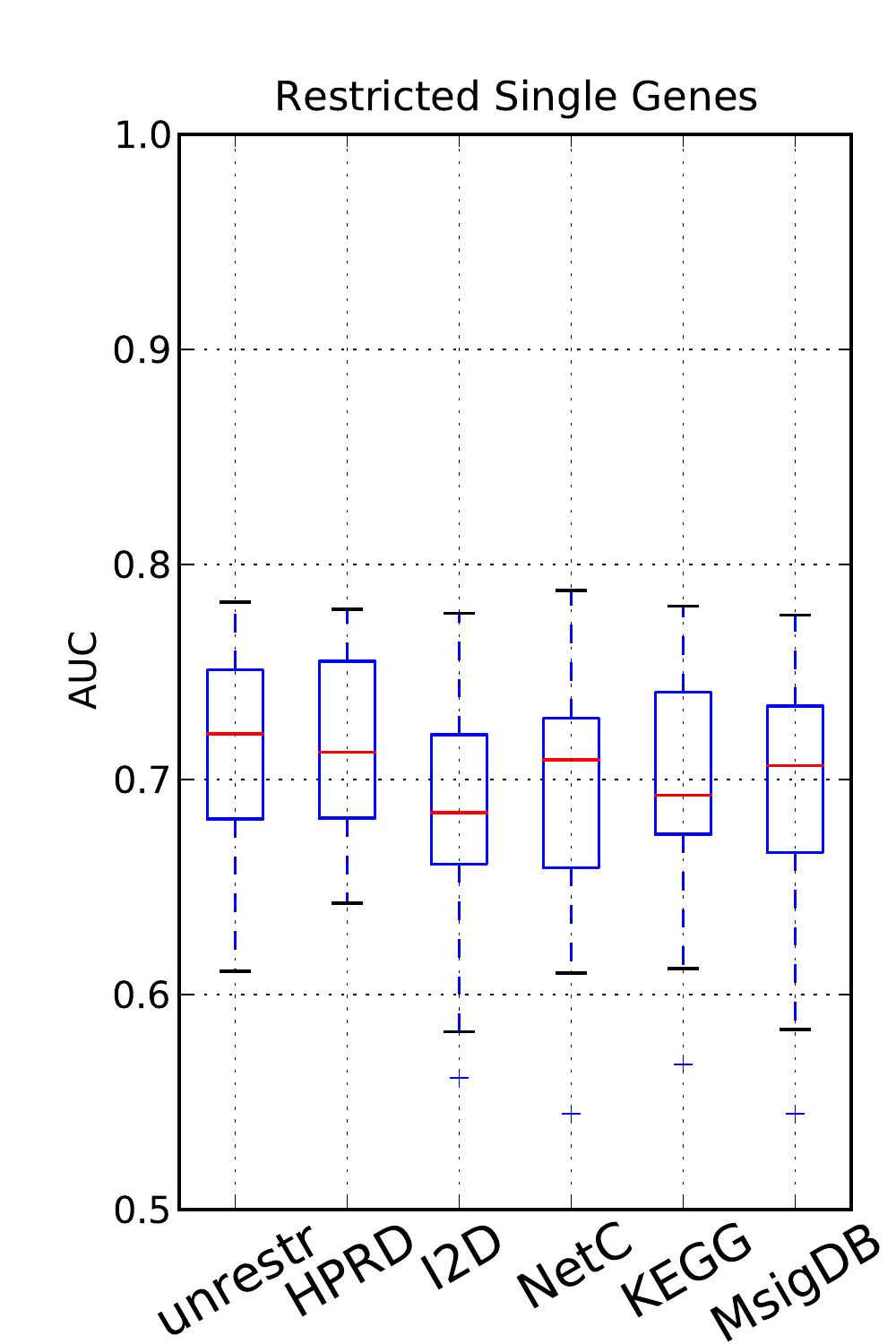}
\end{minipage} 
\begin{minipage}{0.3\linewidth}
B\\
\includegraphics[width=\linewidth]{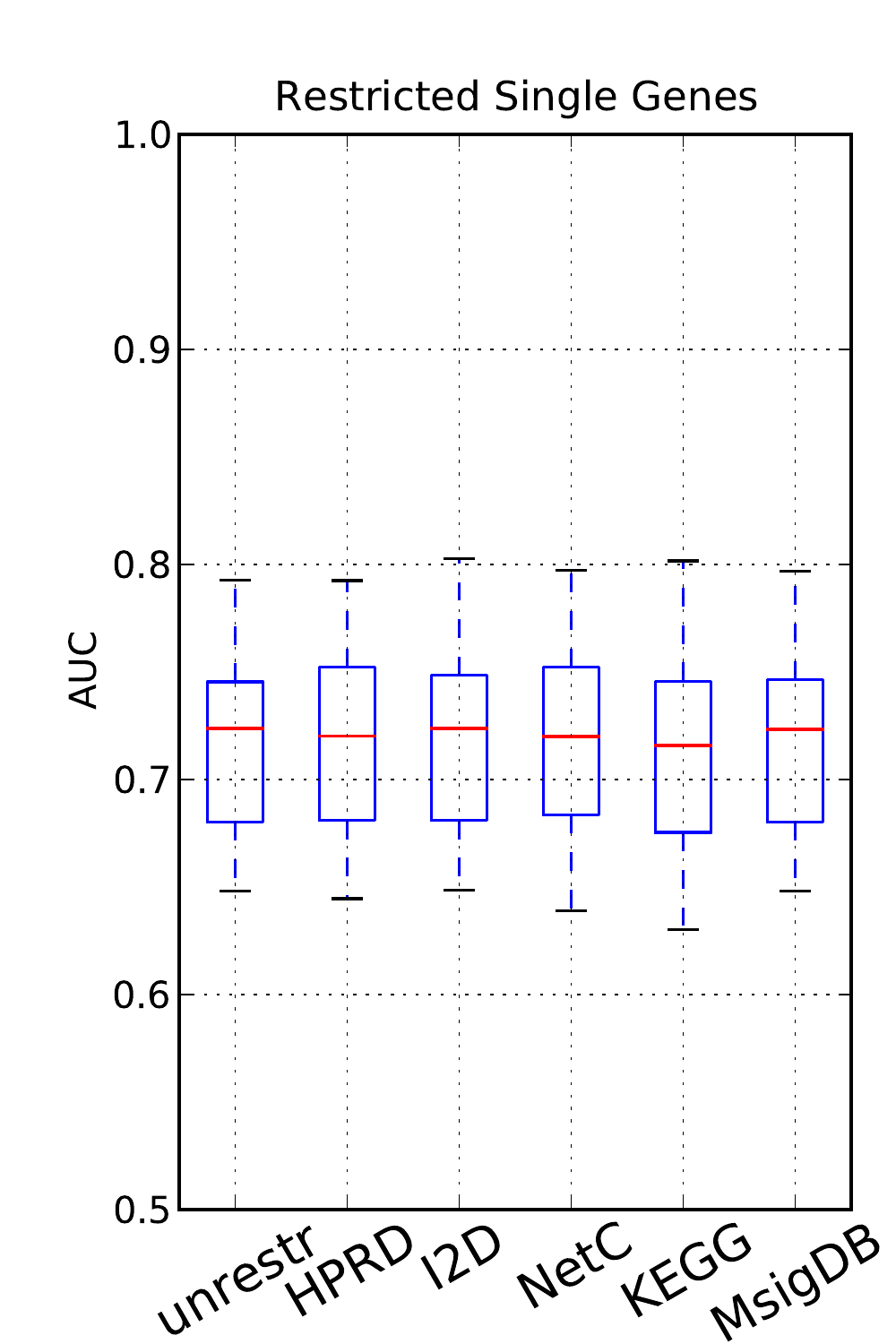}
\end{minipage}
\caption{\small{\bf Comparison of single gene classifiers restricted to genes occurring in the secondary data sources.} 
We compared the performance of the single gene classifier trained on all genes present on the microarray ({\em unrestr.}) with 
the performance of single gene classifiers that only employ genes present in the secondary data sources. 
The resulting classifiers are indicated by the secondary data source whose gene set was employed to train the 
classifier. {\bf A:} The number of single genes was determined during
the cross-validation procedure; {\bf B:} the 50 best scoring single genes were employed.}
\label{fig:restricted_genes}
\end{figure}

We next hypothesized that the lack of difference in the performance between composite classifiers and single gene classifiers 
could be caused by the fact that the composite features are bound to the genes annotated in the secondary data  while single gene 
classifiers can employ all genes on the microarray. To test this 
hypothesis we retrained the single gene classifier, but restricted the set of genes from which features for the final classifier could be 
selected to the genes that are present in the respective secondary data sources.
The resulting classifiers are denoted by the secondary data source from which the gene set is derived, while the single gene classifier employing features from the whole microarray is 
denoted by \textit{unrestr}. The results of this analysis are depicted in Figure~\ref{fig:restricted_genes}A. 
There is significant difference in the performance of the classifiers employing genes annotated in the I2D, 
KEGG and MsigDB (Table S9). However, when accounting for multiple
testing only the difference between \emph{unrestr} and I2D remains
significant. Moreover, as indicated earlier, the optimization of the
number of features by cross-validation introduces significant variation in the number of features without resulting in large performance changes. To eliminate this source of variation from the comparison, we fixed, as before, the number of features to 50, 100 and 150 and repeated the comparisons. The results are depicted in Figures~\ref{fig:restricted_genes}B and S7, and Tables~S10-S12. We can only find significant differences between the
unrestricted single gene classifier and KEGG when the 50 best
features are selected and the I2D when employing 
the 150 best features. However, both of these differences disappear when multiple testing correction is performed. We therefore conclude that the starting gene set has a minor influence on the single gene classifiers. Hence we can reject the hypothesis that feature extraction approaches employing secondary data sources are put at a disadvantage since they can not exploit the full set of genes present on the array.

\subsubsection*{Training set size has no significant effect on performance differences}

A third possible factor that could explain the lack of performance difference between the composite feature classifiers and the single gene 
classifier is the size of the training set. Recall that in the cross-validation procedure we train on one data set and then test on another 
data set. We repeat this procedure for all possible pairs of data sets; excluding, of course, training and testing on the same data set 
(\emph{paired setting}). We can, however, also follow an alternative scheme where we train on \emph{all} data sets except the test set, 
the so-called \emph{merged setting}. More specifically, in this setting four of the five Affymetrix data sets were merged to form a single 
training data set and the fifth data set was used 
as test set. Thus, we receive for each feature selection method five AUC values. This increases the size of the training set, and 
by comparing the results obtained in this setting with the results
from the paired setting, we can investigate the effect of the training set 
size on the classifier performances. 

\begin{figure}[btp]
\begin{tabular}{c|c|c}
 & CV-opt features & 50 best features\\\hline
\begin{sideways} \hspace{2cm} merged\end{sideways} &
\includegraphics[width=3in]{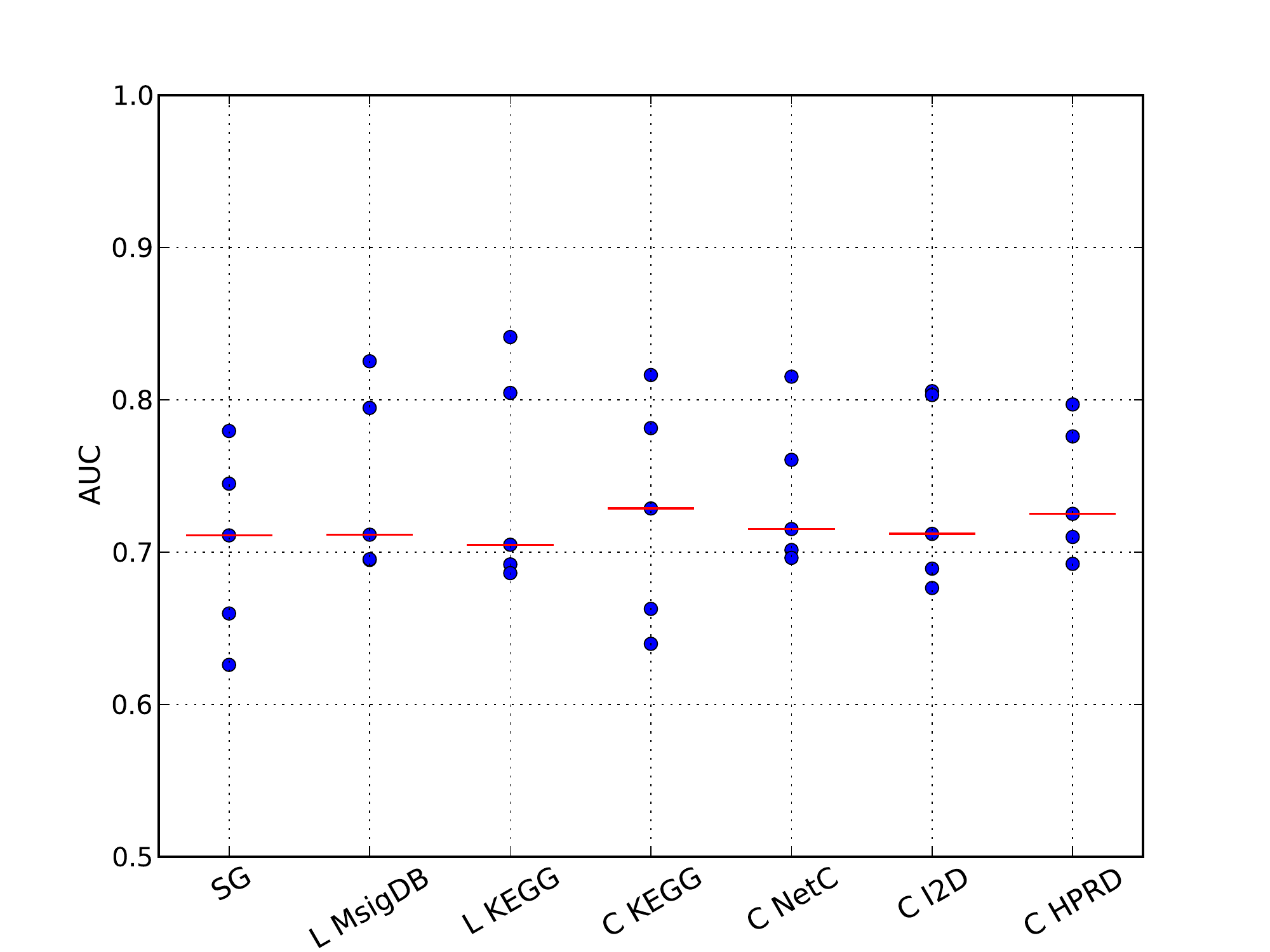} &
\includegraphics[width=3in]{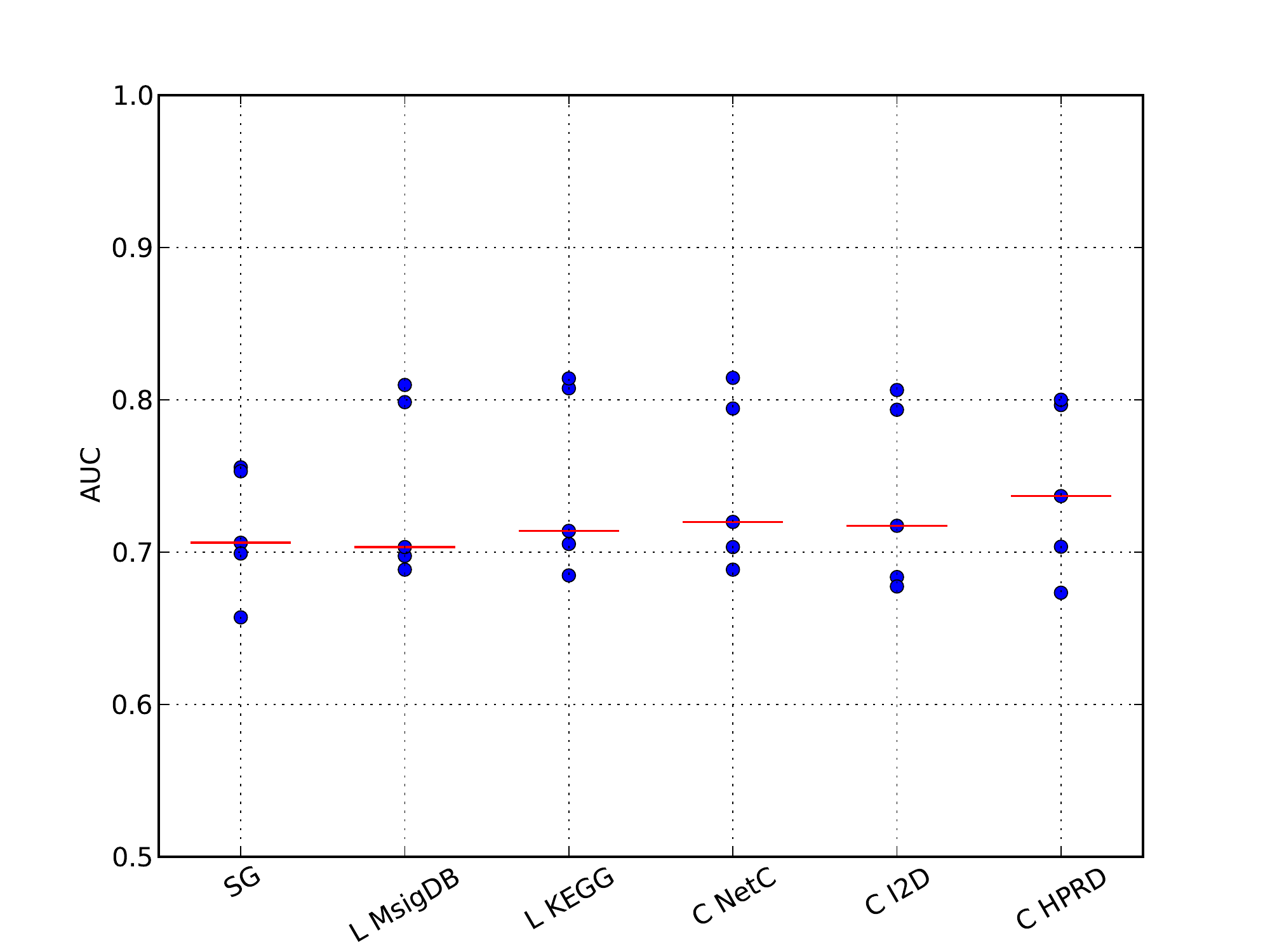} \\\hline
\begin{sideways}\hspace{2cm}pairwise\end{sideways} &
 \includegraphics[width=3in]{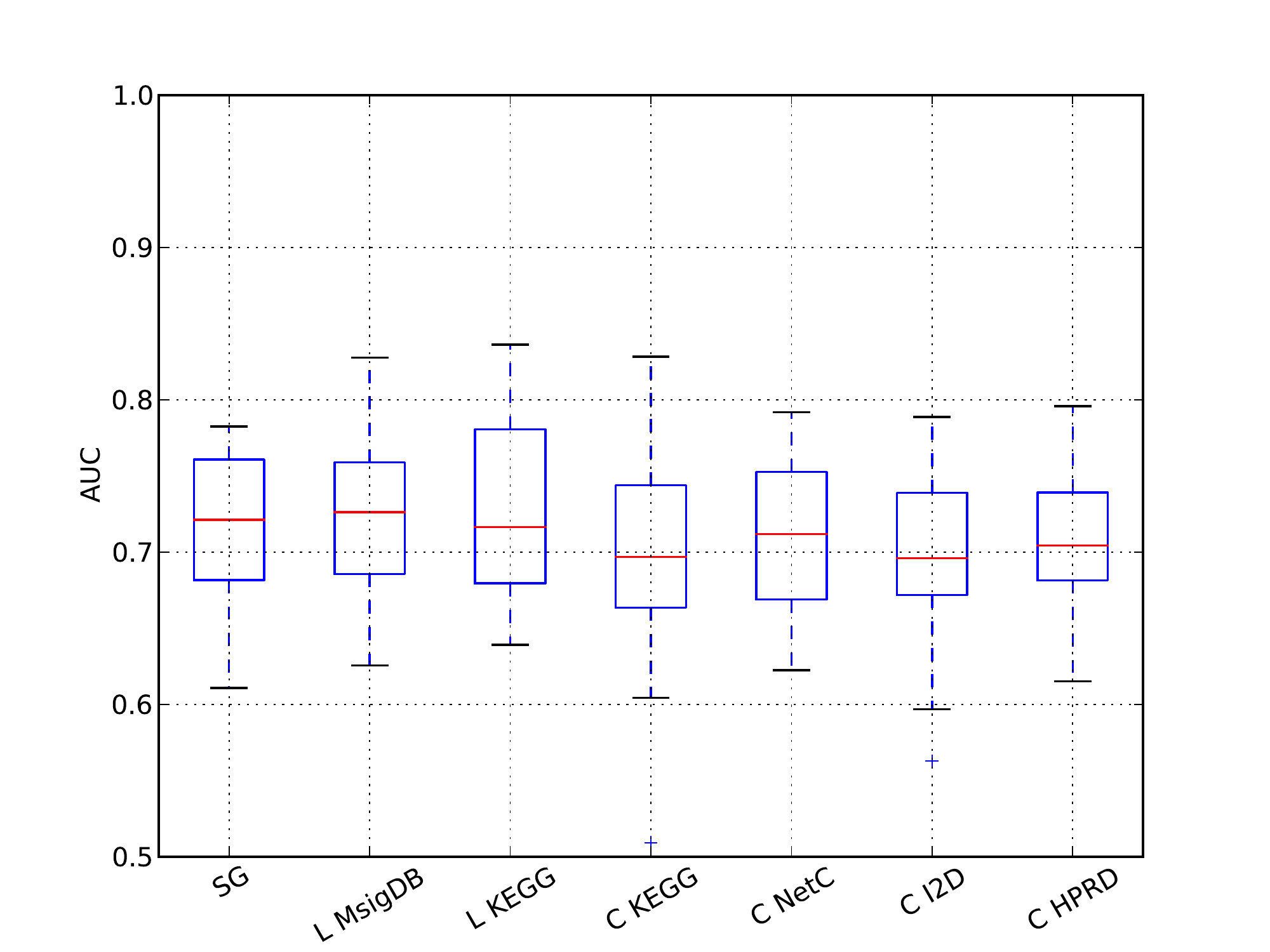} &
 \includegraphics[width=3in]{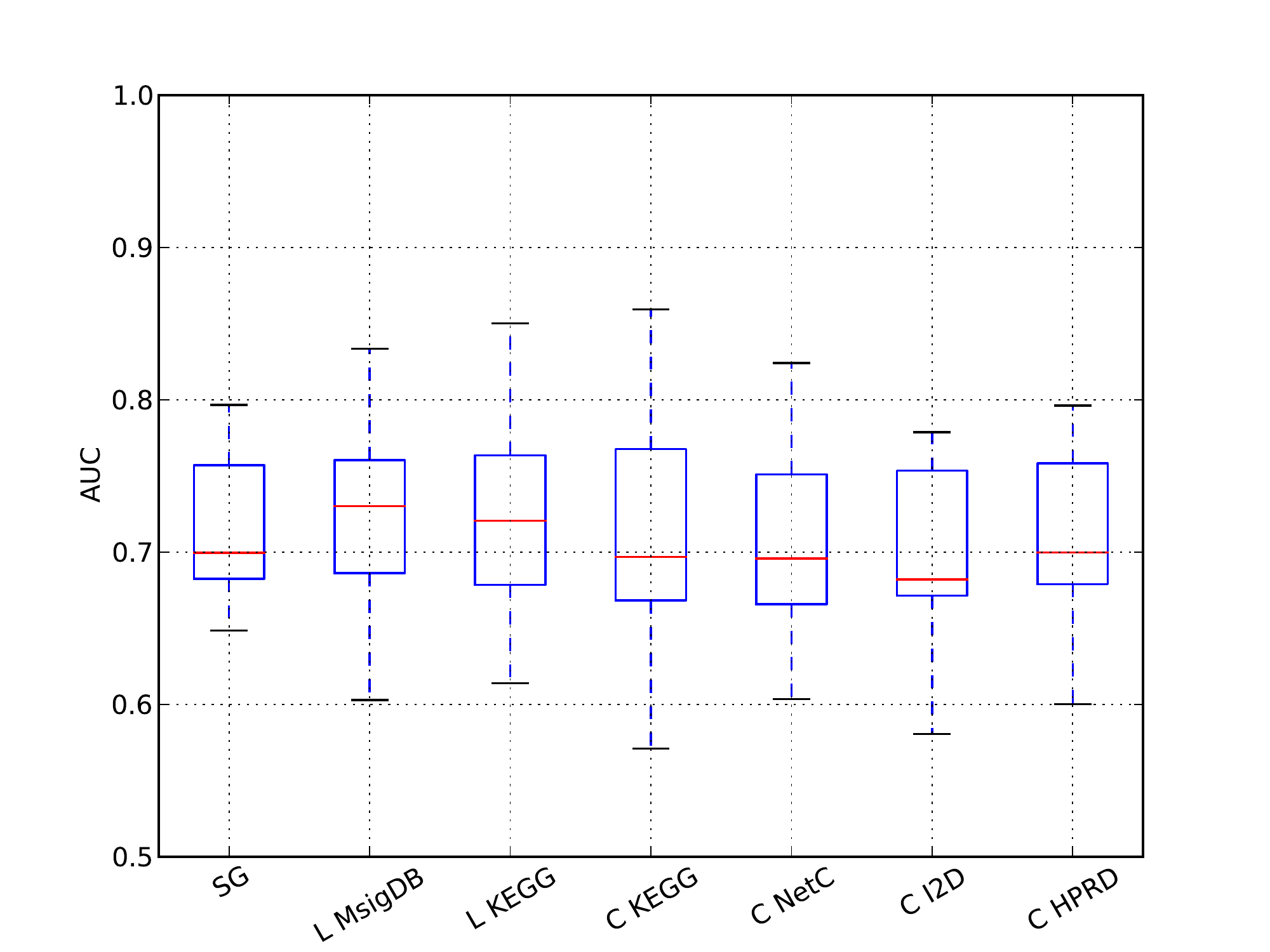}\\
\end{tabular}

\caption{\textbf{Classification results for merged and paired setting}. In the merged setting one Affymetrix data set is set aside 
as test and the remaining four Affymetrix data sets are merged into a single data set. This is repeated until every one of the five data sets 
acted as a test set. {\bf Top row:} Results for the merged setting. The red lines indicate the median. {\bf Bottom row:} Only the five Affymetrix data sets were used in the paired setting.}
\label{fig:boxplot_pw_vs_merged}
\end{figure}

Figure~\ref{fig:boxplot_pw_vs_merged} depicts the results for the
merged setting and the pairwise setting for the CV-optimized feature sets and 
when only the top 50 features are selected. (Note that, in contrast to the results in Figure~\ref{fig:auc_boxplot}, this pairwise setting only employs the Affymetrix datasets). The results for the top 100 and 150 features are similar, see  
Figure S8. Statistical testing shows that in the paired setting (Tables S13-S16) when the number of features is set to 150, \Lee employing the MsigDB performs better than the single gene classifier. However, this difference disappears when correcting for multiple testing.
More importantly, there are no significant differences between the performances of the single gene and composite feature classifiers
in the merged setting (Tables~S17-S20). 

%However, we see small differences in 
%the AUC value distributions of the classifiers in the paired setting. When setting the feature number to 150, \textit{Lee} employing the MsigDB performs 
%better than the single gene classifier. However, the performance of all other composite classifiers is not better than the performance of 
%the single gene classifiers (Tables S13-S20). \todo{g: does this
%  belong here? $\to$ discuss}
%\obacht[Christine]{I did not calculate the Wilcoxon between the pairwise and merged.}
Hence, we can also reject the hypothesis that the lack of performance difference is due to the sizes of the employed training sets. 
%\obacht[lodewyk]{Christine, I am sure you did this many times, but could you double check the single gene results on Loi and Pawitan: 
%why are they not performing as well as networks?}
%\obacht[Christine]{I checked it in the python version and in the Matlab version of our results. Both show the same: For networks 
%Loi and Pawitan show higher AUC than for single genes. I have no explanation for that. I found that both Loi and Pawitan have appr. 80\% of
%ERpos patients. I assume that the networks pick that up. Could it be that 
%single genes are less sensitive to dataset composition?}
% shows that the median performance is the same for both settings (Figure \ref{fig:boxplot_pw_vs_merged}). We find that classifiers using the Loi or the Pawitan data set as testing data set have in general a higher performance in the 'merged setting' using \Lee or \Chuang features (Figure \ref{fig:pw_vs_merged}). However, a similar phenomen is observed in the paired setting; classifiers trained on the Loi data set perform best when tested on the Pawitan data set and vice versa (data not shown). These artifacts originate from the higher proportion of ER positive patients (tested by immuno blots) in these two data sets; the Loi and Pawitan data sets contain ~80\% whereas the other data sets contain around 70\% ER positive patients.

\subsubsection*{Dataset homogeneity affects single genes and composite classifiers similarly}

\begin{figure}[tbp]
\begin{minipage}{0.5\linewidth}
\textbf{A}\\
\includegraphics[width=3.5in]{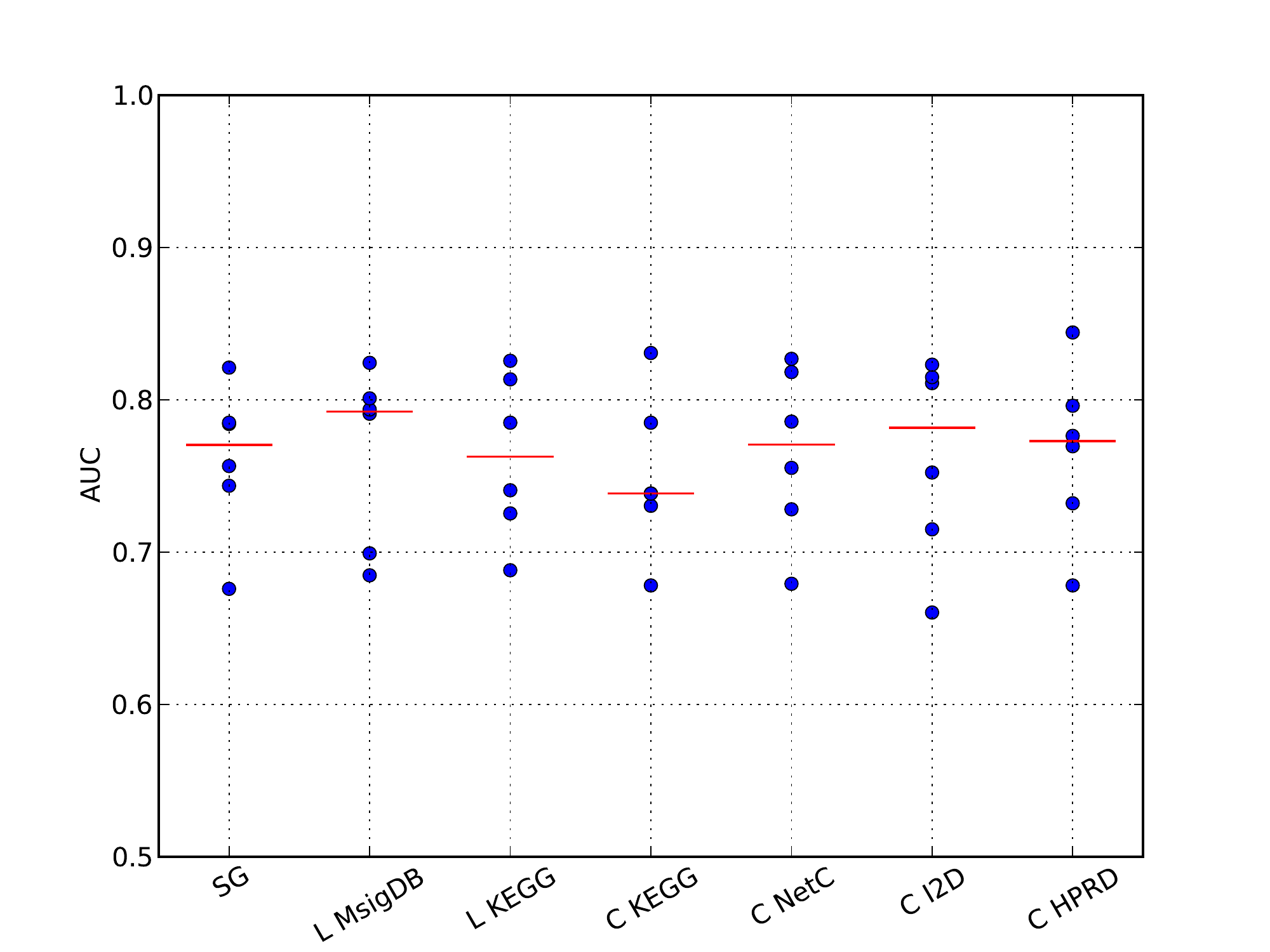} 
\end{minipage}
\begin{minipage}{0.5\linewidth}
\textbf{B}\\
\includegraphics[width=3.5in]{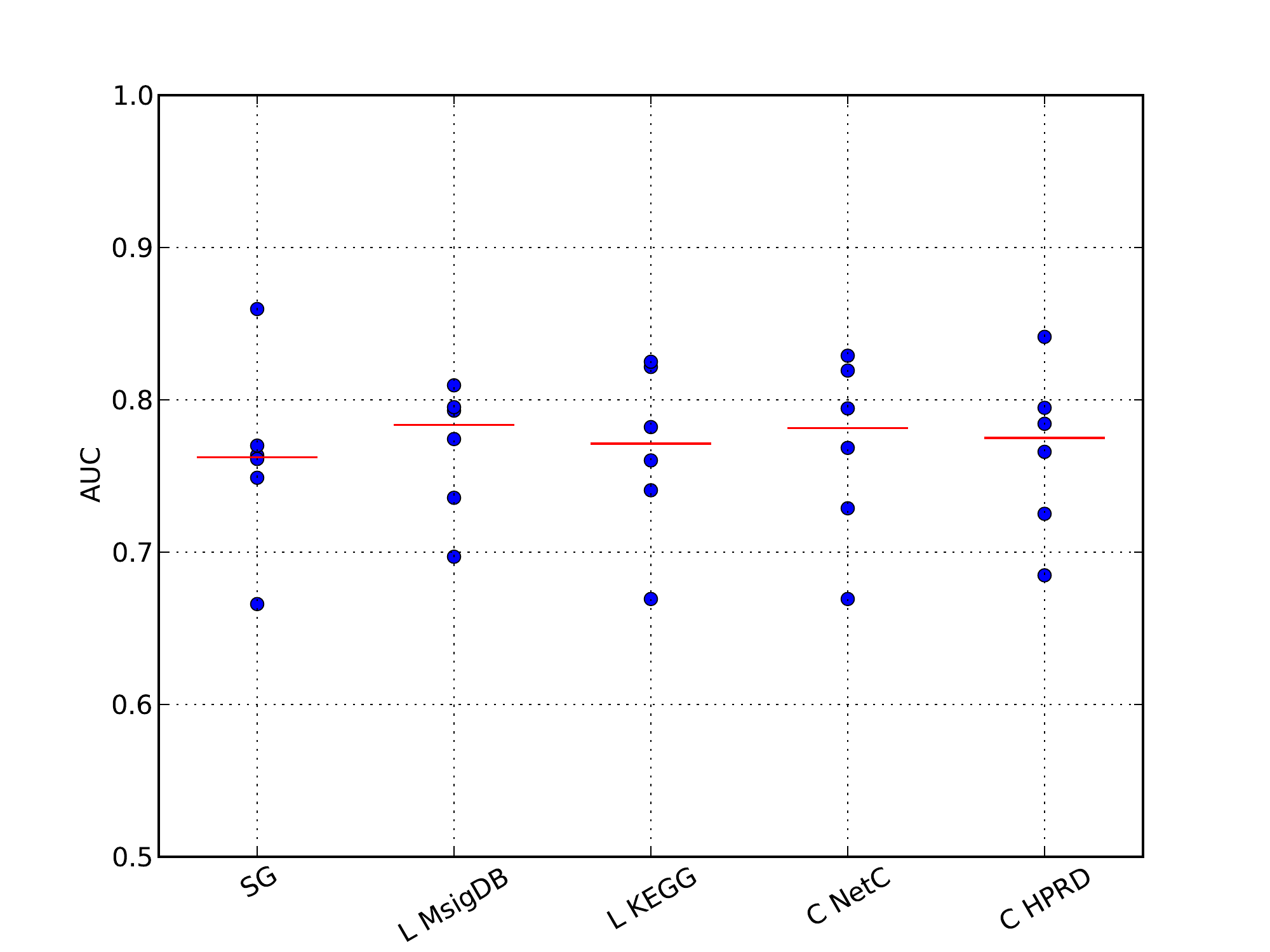} 
\end{minipage}
\caption{\small{\bf Classification results of the ER positive data only.} The ER positive cases from a single data set were set aside as test set while ER positive cases from the remaining 
five data sets were merged into a single training set. This was repeated until each data set was employed as left-out test set, resulting in six 
AUC values. The red lines indicate the median. {\bf A}: CV-optimized number of features; {\bf B}: 50 best features.}
\label{fig:ER_pos}
\end{figure}

Breast cancer is a collection of several heterogeneous diseases that show very different gene expression patterns \cite{Gatza2010}. 
Expression patterns predictive of outcome might vary between
subtypes, which typically leads to problems when training classifiers on gene expression data derived from breast tumors. If this is not explicitly taken into account 
during classifier training it could result in poor performance and unstable classification, as the selected genes may depend on the composition 
of the training set. In this section we control the heterogeneity in both the training and test sets by only selecting the relatively 
homogeneous ER positive breast cancer sub-population. Since the
training sets become too small in the paired setting if we only select the 
ER positive cases, we followed the merged setting outlined above. More specifically, we created a test set consisting of all ER positive 
cases of a single data set and a training set by pooling all ER positive cases from the remaining data sets. 
%\obacht[lodewyk]{Did we only pool the Affy ER+ cases? It states in
%the text 6 data sets.} 
We repeated this procedure across the six data 
%\obacht[Christine]{I pooled here all 6 datasets}
sets and thus obtained six AUC values per classifier. Figure~\ref{fig:ER_pos} depicts the results for the CV-optimized feature sets and the top 
50 features. As before, the classifiers employing a fixed number of features perform similar to classifiers based on a feature set optimized 
in the CV procedure. See Figure~\ref{fig:ER_pos}B and Figure S9. In general, and in accordance with earlier 
observations as made, \eg, by Popovici \ea \cite{Popovici2010}, the performance of all classifiers is substantially 
better on the ER positive cases compared to the unstratified case. More importantly, also in this setting there are no significant performance 
differences between the single gene classifiers and composite feature classifiers (Tables S21-S24). 

$$$$$$$$

%%%%%%%%%%%%%%%%%%%%%%%%%%%%%%%%%%%%%%%%
%\todo{What is preferable:
%\begin{itemize}
%\item{First ER stratification (by pooling ER+ in Affy and Agilent) then followed by effect of pooling all samples?}
%\item{First Pooling (of Affy only) then followed by ER+ stratification}
%\item{Problem: we pool ER+ across platforms, then we only pool Affy: question: why did we not also only pool the Affy ER+?}
%\item{Alternative: never pool Affy only: neither in ER+ not in the complete data sets}
%\end{itemize}}
%%%%%%%%%%%%%%%%%%%%%%%%%%%%%%%%%%%%%%%%

\subsection*{Equal classification using real or randomized networks and pathways}

\begin{figure}[tbp]
\centering
\includegraphics[width=2in]{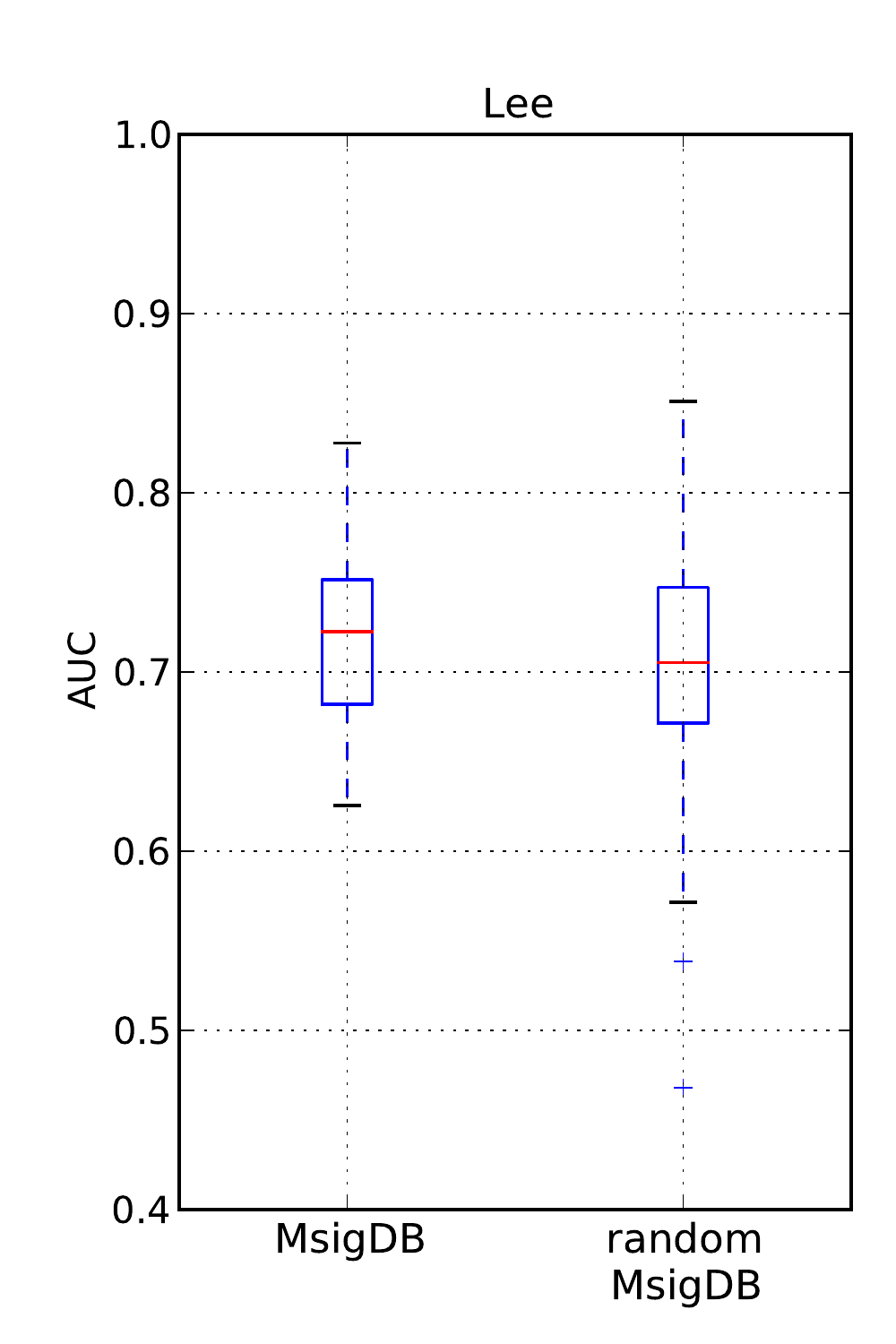}
\includegraphics[width=2in]{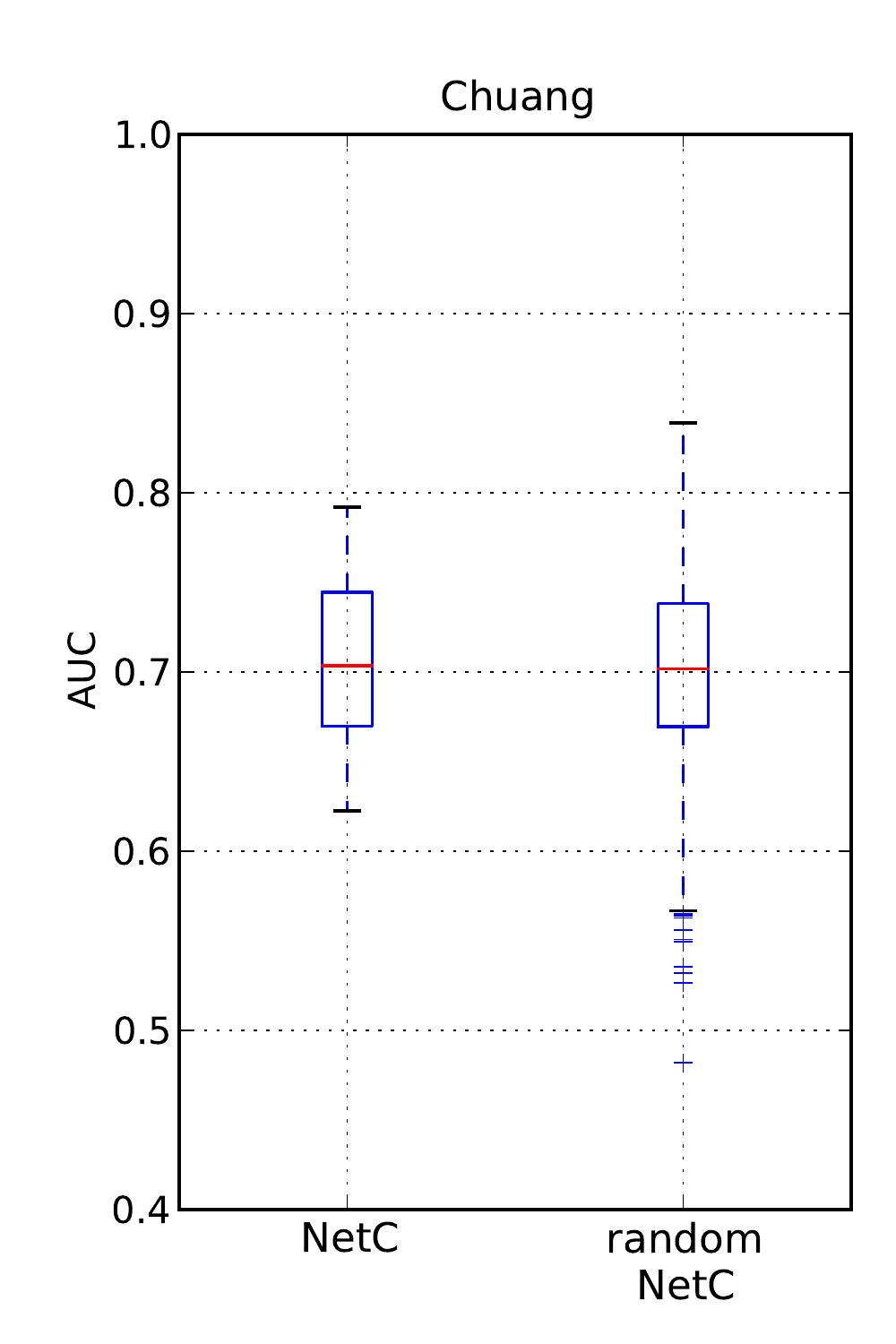} 
\caption{\small{\bf The effect of randomized secondary data sources}. {\bf Left:} AUC values obtained with the feature extraction method \textit{Lee} on real and randomized MsigDB pathways. {\bf Right:} AUC values obtained with the feature extraction method \textit{Chuang} on real and randomized PPI networks.}
\label{fig:rand_vs_pws}
\end{figure}

In the previous sections we showed that composite classifiers do not perform significantly better than classifiers 
employing single genes as features. We investigated several factors that could influence the performances of these two approaches, but 
failed to find any factor that induces significant performance differences on the data sets we employ in this study. This lead us to question 
whether prior knowledge sources really contain information that is of value in constructing features for classifiers predicting outcome in 
breast cancer. Chuang \ea \cite{Chuang2007} compared their PPI-based classifier to classifiers derived from randomized PPI 
networks. They concluded that their classifier performed significantly better than random classifiers. We decided to repeat this analysis for 
a subset of the classifiers in our comparison to determine whether prior knowledge sources really contain information relevant for predicting 
outcome in breast cancer. To this end, we generated, for each prior knowledge source, 25 random instances. More specifically, 
we maintained the structure of the pathways, networks and gene sets, and randomly permuted the identities of the genes. In doing so, the 
original topology of the secondary data is preserved while the biological information is destroyed. We then repeated the whole validation 
procedure on all 25 random instances for the feature extraction methods \Lee and  \Chuang. The results of this analysis are 
presented in Figure~\ref{fig:rand_vs_pws}. Strikingly, classifiers derived from secondary data sources suffer no significant performance 
degradation when employing randomized secondary data sources. The performance of \textit{Chuang} on randomized PPI data clearly has a 
large variance, and there are instances of classifiers derived from random networks that perform much worse and much better than classifiers 
derived from the non-randomized networks. Furthermore, we found that most classifiers based on randomized secondary data show 
performances similar to the classifiers derived from the real secondary sources. To formalize this observation, we performed a statistical test. We have reason to believe that the results derived from the real data should be better than the results derived from random data. Hence we performed one-sided paired Wilcoxon rank tests to determine whether the null hypothesis that the mean `real' AUC-value is larger than each of the the 25 `randomized' mean AUC-values can be rejected. We performed a Bonferroni correction to account for multiple testing. The results in Figures~S10-S12 and Tables~S25-S30 show that in the vast majority of the cases the null hypothesis can not be rejected. Conversely, it is very simple to generate a randomized secondary data source that performs equally well as the real data source. This result shows that further research must be done on the utility of 
secondary data sources in predicting breast cancer outcome.
%This result casts serious doubt on the utility of secondary data sources to improve the prediction performance of prognostic predictors in breast cancer. 

\subsection*{Current composite feature classifiers do not increase the stability of gene markers}
Apart from performance improvements, it is also frequently claimed that features derived from classifiers employing secondary data sources 
are far more stable than single gene classifiers. In other words, whereas single gene signatures extracted from different data sets show very 
little overlap, features extracted by approaches that employ secondary data sources are claimed to show a large degree of overlap, even though 
the features were derived from separate data sets. 

\begin{figure}
\centering
\includegraphics[width=3.5in]{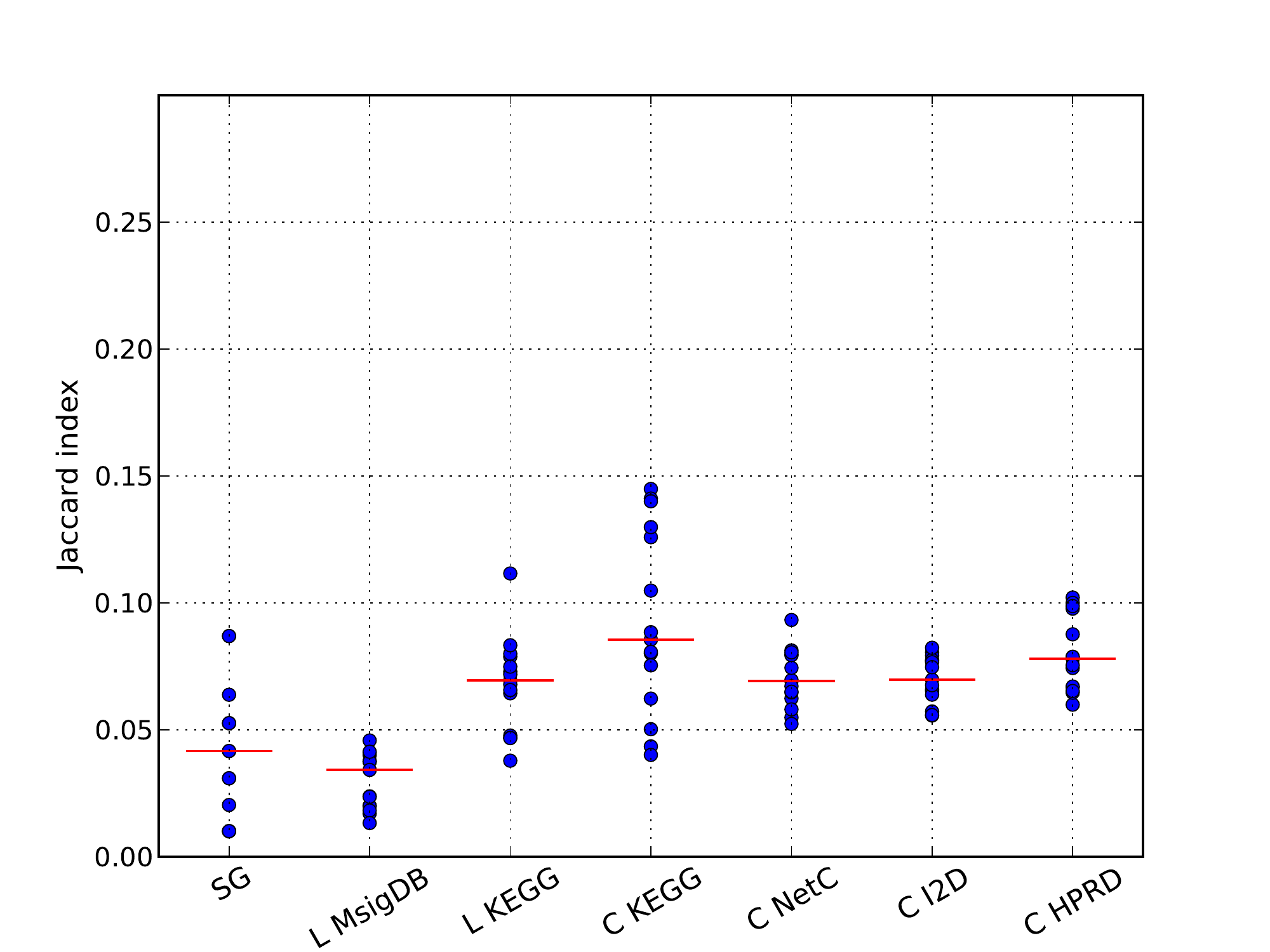} 
\caption{\small
{\bf Feature stability when the top 50 features are selected.} For each method the Jaccard index was calculated between the gene sets 
extracted from two different data sets. This was repeated for all pairwise combinations of data sets. Thus, 15 values were obtained.}
\label{fig:overlap}
\end{figure}

\begin{figure}
\centering
\includegraphics[width=5in]{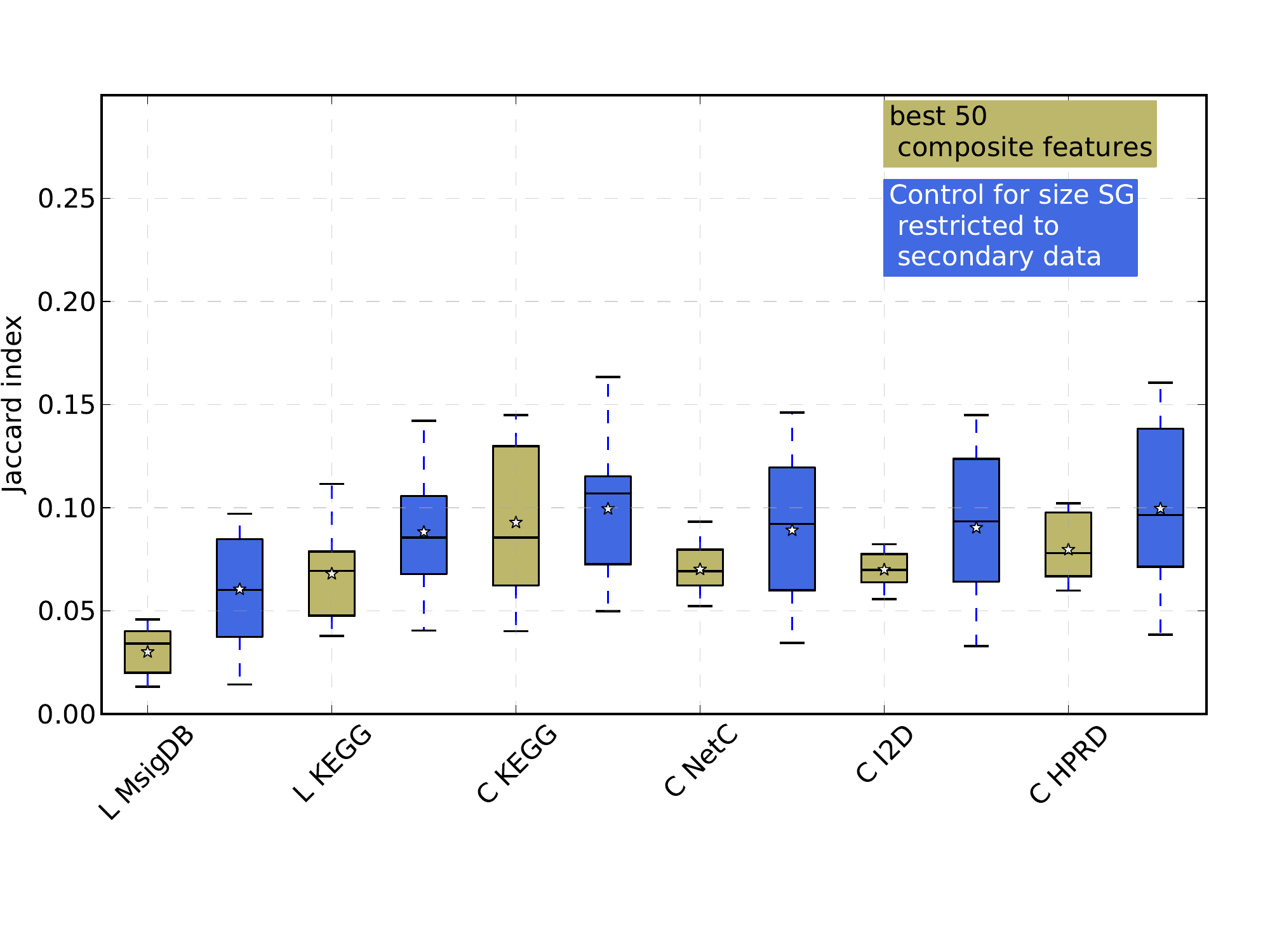} 
\caption{\small
{\bf Feature stability when corrected for gene set size}. Box plots of the Jaccard indices computed for all pairs of gene sets derived from 
two different data sets. The green box plots
represent the Jaccard indices for genes constituting composite
features, while the blue box plots (denoted as `Control for size SG')
represent the gene-size-corrected Jaccard indices for single gene
classifiers. The white stars represent the mean of the
distributions.}
\label{fig:overlap_featuresNMC_cfs}
\end{figure}

In this section we determined whether feature sets extracted from secondary data sources do, 
in fact, show a larger degree of stability than single gene feature sets. %OBACHT gunnar: this sentence didn't make any sense to me and I didn;t feel it is needed -->< I deleted it: We assessed the stability of the single gene feature sets constructed by optimizing the classifier performance in the cross-validation procedure and employed in the classifiers reported on in the first results subsection. 
As a measure of stability we calculated the pairwise Jaccard index between the features derived from different data sets for a given 
feature extraction method. Since the number of features determined in the CV procedure varies, we performed this comparison for the cases where the top 50, 100 and 150 features are selected. 
%comparing these features would introduce another bias.
The Jaccard indices for the best 50 features are depicted in Figures~\ref{fig:overlap} while Figure S13 depicts the results for the best 100 and 150 features. It is clear that the overlap of feature sets 
consisting of single genes is relatively low, albeit slightly higher than the overlap of \Lee-MsigDB\@. The highest consistent stability is obtained by \Chuang-HPRD with \Chuang-KEGG showing high variance in stability.
% employing NetC while all other approaches show intermediate 
%stability. Notably, \textit{Chuang} employing the I2D displays a high variance. Figure \ref{fig:overlap}B depicts the signature stability when only the top 
%50 features are considered. Also in this comparison, single genes are most unstable across data sets, with \textit{Chuang} showing generally 
%high stability.
One can therefore clearly conclude that, when compared in terms of \emph{signature genes} overlap, single genes are generally less stable than 
feature sets 
extracted by including secondary data sources. However, strictly speaking, such a comparison compares the proverbial apples and oranges, 
since a single feature constructed based on secondary data sources can contain many genes. In order to ensure that the low overlap of single 
genes is not only due to the fact that the best single gene feature sets contain fewer genes than the other feature sets, we controlled the 
single gene feature sets for size. More specifically, for each data set and each feature selection approach employing secondary data sources, 
we obtain a single best feature set consisting of $n^*$ features (networks, gene sets or pathways) that, in turn, consists of $m$ genes. We then determine a size-matched 
single gene set  by choosing the best $m$ single genes on that same expression data set. We also required 
these single genes to be annotated in the respective secondary data source. For these size-matched single gene feature sets, we 
computed the Jaccard index. The results, depicted in Figure~\ref{fig:overlap_featuresNMC_cfs} and S14, show that when this size correction is applied, 
the stability of single gene feature sets are as high as features
extracted by employing secondary data sources. See also Tables~S31-S33.

%\textcolor{green}{As a matter of fact network and pathway-based features usually contain more than genes than involved in a single gene 
%classifier, e.g.\ 50 network or pathway features can contain up to 500 genes. In order to account for that we fix the number of single genes 
%to the number of genes involved in the pathway and network-based features.} 

\section*{Discussion}

In this study we evaluated the prediction performance of network and
pathway-based features on six breast cancer data sets.  In contrast to
previous studies we found that none of the classifiers employing
composite features derived from secondary data sources can outperform
a simple single gene classifier. Moreover, we did not find any
evidence that composite features show a higher stability across the
six breast cancer cohorts. Our findings suggest that with the
feature extraction methods tested in this study, we cannot extract
more knowledge from secondary data sources than we find in the
expression of single genes.

There are several issues that could potentially contribute to that
situation. First, secondary data sources are, to a large degree,
generated by high-throughput biological experiments and could thus contain a  level of noise that deems them inappropriate for outcome prediction in breast cancer. On the other hand, the search algorithms could be
unsuitable to detect biologically meaningful networks. All
three feature extraction methods only extract local information
without taking into account the full structure of the network or
pathway data. This local information is then combined in the
classifiers in a rather crude way, namely by simply averaging the expression of the genes associated with a feature that was found to be associated with outcome, \ie
treating each single sub-network or sub-pathway as a single dimension in the classification
space. Possible dependencies between features are not taken into
account. Also, exploring the subnetwork search space in a heuristic
manner may decrease classification performance. The recent method by Dittrich \ea \cite{Dittrich2008}
computes provably maximally deregulated connected subnetworks based on
a sound statistical score definition. This method has not yet been used for classification. 

Other recent algorithms as presented by
Ulitsky \ea \cite{Ulitsky2010}, Chowdhury \ea \cite{Chowdhury2011} and
Dao \ea \cite{Dao:2010kt}
search for deregulated subnetworks in subsets of samples. 
These subnetworks are sets of genes that are
deregulated in most, but not necessarily all, patients with poor disease
outcome. The heuristic method by Chowdhury \ea \cite{Chowdhury2011} has been shown to perform
well on cross-platform classification of colorectal cancer
outcome. Dao \ea \cite{Dao:2010kt} improved on these results by
exact enumeration of all dense subnetworks with the above-mentioned
property. Looking at subsets of samples in a class, \ie a subset of the poor outcome samples, is an interesting
aspect for further evaluation, in particular for breast cancer outcome prediction as it may
quite accurately capture the large phenotypic variety of this rather
inhomogeneous disease. 

%%obacht g: don't discuss the recent paper by dao et al. here
%%\cite{Dao:2011fu}. this is indeed a very interesting paper. 

% \todo{l: I do not see the link of this statement to the rest of the paragraph: `A general rule in classification states that the number of features should not be much larger than the number of samples to allow for a good 
% abstraction from the sample values.
% Lee \ea \cite{Lee2008}  and Chuang \ea
% \cite{Chuang2007} achieve such a low number of features without
% loosing information of single gene expression by using a network score
% that is a variant of the average expression of the genes
% included.} \\
%\todo{l: I would suggest:}
Lee \ea \cite{Lee2008}  and Chuang \ea \cite{Chuang2007} average the expression values of single genes comprising a subnetwork, to determine the `activity' of the subnetwork. This is, however, a very simplistic view of the dynamics 
in a subnetwork itself. In contrast to these two algorithms, the
method by Taylor \ea \cite{Taylor2009} predicts outcome by trying to capture the disruption of the regulation of a hub protein over its interactors in poor outcome patients. This is implemented by using every 
edge leading to a hub as a feature in classification space. Yet, in this way, the classification space becomes too large to 
allow for good classification results. This method is thus not appropriate for solving the classification problem and this is clearly demonstrated in the poor performance of this algorithm in the comparison.

To find a subnetwork scoring function remains one of the biggest problems when including promising
gene sets into a classification framework. Abraham \ea \cite{Abraham2010} tested the classification performance of 
gene sets provided by the MsigDB\@. The authors employed several set statistics like mean, median and first principle component to score the 
gene sets. They found that none of the classifiers employing gene sets and scoring them with the set statistics performed better than 
a single gene classifier.

%Another source of noise is the gene expression data. \obacht[l]{This is the same for single genes and networks}
In our experiments where we shuffle the genes in the secondary data we showed 
that features determined on this nonsensical biological data perform equally well in classification than features determined on the real 
secondary data. This again could possibly be caused by the low quality of the network and pathway data. However, the nature of gene expression patterns in breast cancer, and specifically its association with outcome can also explain these findings. Since many genes are involved in breast cancer and are  
differentially expressed and associated with outcome, as shown, for
example by Ein-Dor \ea \cite{Ein-Dor2005}, the chance that those genes span decently sized subnetworks in the randomized secondary data is high. Both algorithms, \Chuang and \Lee, look for highly differentially expressed subnetwork or 
pathway markers, and these can also be found in the randomized data. 
%It has already been shown by Ein-Dor \ea \cite{Ein-Dor2005} that there are many differentially expressed genes, which, in different combinations, lead to the same classification performance.
Furthermore, overlaying networks or pathways that contain protein level information with mRNA expression data might result in 
erroneous results. These data sources reflect events on very different molecular levels. While gene expression and protein expression is undoubtedly linked, there are many processes that prevent this from being a trivial one-to-one mapping.
%influence the synthesis from mRNA to fully functioning proteins, such as, \eg degradation of malfunctioning mRNA or secondary protein modifications. 
Thus, we may measure, for a set of genes, an effect on the mRNA level that leads to differential expression between 
the two patient classes but this may have little bearing on the relationships between these genes as captured in the PPI graph.
%ano effect on the cellular level like,
%for example, abnormal growth of the cell. Even worse, we can also 
%measure no significantly changed mRNA expression levels whose corresponding proteins will be malfunctioning when secondary 
%modifications, for instance, phosphorylation, are disabled.
In conclusion, our results show that it may not be sufficient to search for sub-networks or sub-pathways that are differentially expressed \emph{on average} but that complex interactions between entities as well as the more complicated relationships between mRNA levels and the topology of the PPI graph need to be taken into account.
%secondary data. %%\obacht[gunnar]{but to do what instead?} 

%\todo{g: I find the following topic/paragraph too marginal to be presented in
%  this prominent length and position.} None of the previous studies \cite{Chuang2007, Lee2008, Taylor2009, Ulitsky2010, Chowdhury2011} gives explicit hints how to treat 
%proteins in the network for which we have no expression data. The handling of this issue is very specific to the network search. In 
%algorithms where given gene or protein sets, like in the pathway data, are searched for a combination of their members that 
%yield a high differential score, such genes can be easily discarded without changing the biological meaning of the pathways. In 
%network searches it is rather detrimental when genes or proteins are excluded from the network, since this disrupts the network 
%and links between important parts of the network might be destroyed.

Our different classification results are partially owed to the fact that we used a different cross-validation procedure, which, in 
our opinion, fits the clinical situation, better.
The studies by Chuang \ea \cite{Chuang2007}  and Lee \ea \cite{Lee2008} also determined possible features on one data set. 
However, in contrast to our work they reranked the features on the second (test) data set. Furthermore, they determined the number of 
features and the classification performance on this second data set
using five-fold cross-validation. In our opinion, this does not represent a \emph{bona fide} independent validation of the classifier.
% One can thus not speak of a `between' data set evaluation. \todo{too harsh?}
% The studies by Chuang \ea \cite{Chuang2007}  and Lee \ea \cite{Lee2008} employ a cross-validation in which a feature set is 
% already given and seen as a fixed parameter. Thus, their study rather shows the prediction of these specific set of subnetworks or gene sets across data sets.
% Rather than testing the benefits of such a fixed set of subnetworks or gene sets, we aimed at showing the superiority of a certain 
% feature extraction algorithm. 
% \todo{g: i find this confusing. we
%   should certainly mention the different CV procedures. but this way I
%   don't get it $\to$ discuss}

In summary, we introduced a framework to test the use of feature extraction methods with respect to the prediction of their determined
features. We used this framework to specifically test the superiority of feature extraction methods based on network and pathway data 
over classifiers employing single genes. Across six breast cancer cohorts, we showed that the three tested methods do not outperform 
the single gene classifier nor do they provide more stable gene signatures for breast cancer. 

An important aspect that hampers progress in the field of network and
pathway based classification is the lack of proper evaluation of
proposed algorithms. In our opinion this is caused by (1) lack of
reproducibility of the results; (2) lack of large and standardized
benchmark sets to test proposed algorithms and (3) lack of a
standardized, unbiased protocol to assess the performances of proposed
methods on the benchmark sets. To overcome these issues, we have
created a software pipeline that implements all the classifiers as
faithfully as possible and also runs our validation protocol. We have
also established a large collection of breast cancer datasets (and
this is currently being expanded) on which the algorithms can be
tested. Both the datasets and the pipeline are freely available. In
the long term, we envision a web service where a classifier can be
submitted as a software package. The server will then autonomously
evaluate the performance of the classifier using the standardized pipeline on the benchmark set.

\section*{Materials and Methods}
\subsection*{Microarray data}
The microarray data sets used in this work is listed in Table~\ref{tab:data_sets}.
To combine the five Affymetrix arrays with the Agilent arrays we first matched the probes on the arrays to Entrez GeneIDs. Only those genes were 
included in the feature sets that appeared on both platforms, resulting in 11601 genes in total. In case that several probes on one chip matched to the same
gene the expression values of the probe with the highest variance was taken. The final expression matrices were then z-normalized such 
that the expression distribution of each gene has a mean of zero and a standard deviation of one. Samples in the data sets were labeled `good' outcome if 
no event, that is, a distant metastasis or death, occurred within five years. Otherwise samples were labeled `poor' outcome.

\begin{table}[tbp]\small
\caption{%%
\bf{Microarray expression data}}
\centering
	\begin{tabular}{llrl}
	Dataset  &  Outcome  &  poor/good\ &  platform\\ \hline
	Chin \cite{Chin2006}  &  Metastasis within 5 years  &  68/29  &  Affymetrix\\ 
	Desmedt \cite{Desmedt2007}  &  Metastasis within 5 years  &  91/29  &  Affymetrix\\ 
	Loi \cite{Loi2007}  &  Metastasis within 5 years  &  92/28  &  Affymetrix\\ 
	Miller \cite{Miller2005}  &  Death within 5 years  &  156/37  &  Affymetrix\\ 
	Pawitan \cite{Pawitan2005}  &  Death within 5 years  &  120/22  &  Affymetrix\\ 
	Vijver \cite{Vijver2002}  &  Metastasis within 5 years  &  178/70  &  Agilent\\ 
	\end{tabular}
\begin{flushleft}
\small
{\bf
Expression data used in this study.} All data sets were processed as described in \cite{Vliet2008} and contained 11601 genes with 
z-normalized expression values afterwards. Column `poor/good' contains the number of samples with poor or good outcome, respectively.
\end{flushleft}
\label{tab:data_sets}
\end{table}

ER status of patients was predicted from the expression values of the gene ESR1. 
%patients with an expression value greater than $-1.84$ were labeled ER positive. 
For more detail of the processing of the data see van Vliet \ea \cite{Vliet2008}.

%All data sets were processed as described in \cite{Vliet2008} and contained 11601 gene probes that were matched to Entrez GeneIDs. The data was z-normalized and patients occurring in more than one 
%breast cancer cohort were removed.\par

%In Section ``Influence of the cross-validation procedure and the classifier type on the performance'' we used the original data sets and 
%networks and subnetwork markers from the original study by Chuang \ea \cite{Chuang2007}. \todo{All expression data used in this study are part of our \ldots}

\subsection*{Network and pathway data}
All feature extraction methods were run on the databases KEGG \cite{Kanehisa2010} and HPRD \cite{Mishra2006}. The algorithm by 
Lee \ea \cite{Lee2008} was also run on the MsigDB C2 database \cite{Subramanian2005}.

\subsubsection*{KEGG}
We collected all pathway information available for \emph{Homo sapiens} (hsa) from the KEGG database, version December 2010. 
The entries contained 
information on metabolic pathways, pathways involved in genetic information processing, signal transduction in environmental 
information processing, cellular processes and pathways active in human disease and drug development. We obtained 215 pathways. 
The nodes contained in the 
pathways were matched with the KEGG gene database such that each node carries an Entrez GeneID\@. In this way we obtained a network 
composed of 200 pathways containing 4066 nodes and 29972 interactions of which 3110 nodes are also contained in the expression sets. 
\subsubsection*{MsigDB}
As second pathway database we used the C2 collection of the MsigDB (version 3) \cite{Subramanian2005}, which was also used in Lee \ea \cite{Lee2008} 
(version 1.0). It contains gene sets from online pathway databases such as KEGG, gene sets made available in scientific publications 
and expert knowledge. We obtained 3272 gene sets of which 2714 could be entirely or partially mapped the six data sets. 
The MsigDB does not contain any edges, thus this database was only usable for the algorithm by Lee \ea \cite{Lee2008}.
\subsubsection*{HPRD}
The HPRD (version 9) provides information on protein-protein interactions (PPI) derived from the literature. 
The HPRD contains 9231 proteins and 35853 interactions. The proteins were mapped to their genes carrying Entrez GeneIDs. 
There are 7390 genes contained in both the HPRD and the expression sets.
\subsubsection*{OPHID/I2D}
The OPHID/I2D database, downloaded in April 2011, contains
protein-protein interactions derived from BIND, HPRD and MINT 
%%\obacht[gunnar]{citations needed for BIND, MINT?}, not answered...
as well as predicted 
interactions from yeast, mouse and \textit{C.\ elegans}. The database contains
12643 nodes and 142309 edges. 9453 of the nodes are also present in the six breast cancer studies examined here.

\subsubsection*{Protein-protein interaction network by Chuang \ea \cite{Chuang2007} (NetC)}
The network curated by Chuang \ea \cite{Chuang2007} consists of 57228 interactions and 11203 nodes of which 8572 are contained in
the six breast cancer studies. The network is curated from yeast two hybrid experiments and interactions predicted from co-citation.

\subsection*{Algorithms}
\subsubsection*{Notation}
Let $E_{k\times n}$ be a gene expression matrix, as we obtain it from a microarray study, with $k$ samples and $n$ genes. 
Each entry $e_{i, j}$ contains the expression value of gene $j$ in sample $i$. All samples carry a binary class label 
$l_i \in \{0, 1\}$ denoting outcome, where 1 denotes `poor outcome' and 0 
denotes good outcome'. The label vector of all samples is 
denoted by $L = (l_1, \dots l_k)^T$. We denote a network by $N =(G, I)$ where $G$ is the set of genes in the network and $I$ is 
the set of interactions between 
these genes, also called edges in the following. We define a subnetwork as the connected graph $N^\prime = (G^\prime, I^\prime)$ with
$G^\prime \subseteq G$ and $I^\prime \subseteq I$, 
and a pathway as a gene set $G^\prime \subseteq G$. Let $G^\prime$ be such a pathway or the set of genes in a subnetwork then according 
to \cite{Chuang2007, Lee2008}
the activity of the pathway or subnetwork in sample $i$ is given as
\begin{equation}
\label{activity}
a_{G^\prime, i} = \sum_{j \in G^\prime} \frac{e_{i, j}}{\sqrt{|G^\prime|}}
\end{equation}

\subsubsection*{Feature extraction method by Chuang \ea \cite{Chuang2007}}
Given a network $N = (G, I)$, the algorithm by Chuang \ea \cite{Chuang2007} carries out a greedy search starting from a seed---a single gene in the network. It then iteratively extends the network by adding neighboring genes to find subnetworks with
high mutual information (MI) of the activity of the pathway and sample labels. Each node in $N$ is 
used once as seed. In each step, an additional gene is picked that leads to a maximal MI improvement. If no improvement is possible, the search stops. 

More precisely, the association between the subnetwork activity and the class labels is computed as follows:  
Given a subnetwork $N^\prime = (G^\prime, I^\prime)$ the activity vector $a_{G^\prime} = (a_{G^\prime, 1}, \dots, a_{G^\prime, k})^T$ 
is calculated using 
Equation~(\ref{activity}). To compute the MI, vector $a_{G^\prime}$ is discretized. Given a dissection of the interval $[\min_{i \in [1, \dots, k]} a_{G^\prime, i}, \max_{i \in [1, \dots, k]} 
a_{G^\prime, i}]$ into $\lceil\log_2 k\rceil$ bins let \mbox{$\delta: \mathbb{R} \rightarrow [\delta_1, \dots, \delta_{\lceil\log_2 k\rceil}]$} be a 
function that assigns a network activity to a sample with one of these bins, where $\delta_i$, $i = 1, \dots, \lceil\log_2 k\rceil$, denote the bins.
%and let $b_j = |\{\delta(a_{N^\prime,i}) | i = 1, \dots, p \wedge j = 1, \dots, 9\}|$ denote the number of patients that fall into bin $j$. Then $s_MI$ is defined as follows:
We define the mutual information $s_{\mathit{MI}}$ between the probability density of the bins $p(\delta(\cdot))$ and the probability 
distribution of the class labels $p(l)$ as

\begin{equation}
s_{MI}(\delta, l) = \sum_{i = 1}^{\lceil\log k\rceil} \sum_{l = 0}^1 p(\delta_i, l)\log\frac{p(\delta_i, l)}{p(\delta_i)p(l)}
\end{equation}

where $p(\delta_i, l)$ is the joint distribution of $p(\delta(\cdot))$ and $p(l)$.
The algorithm also performs three statistical tests to extract only networks that show significantly high mutual information. 
The ranking of the networks is given by ordering the networks according to their mutual information $s_{\mathit{MI}}$. 

In our study we use PinnacleZ, an implementation of the algorithm provided by the authors. As feature values for classification the 
subnetwork activity as given in Equation~(\ref{activity}) of the determined subnetworks was used. 

Before determining the subnetworks, PinnacleZ performs a z-normalization of the given data set. This is undesirable when 
looking at subsets of data sets as we do in the five-fold cross-validation. In order to skip the normalization step, 
we implemented a patch in the PinnacleZ source code. This patch adds a ``-z'' option that instructs PinnacleZ to \emph{omit} 
its usual gene-wise z-normalization step.

One problem when mapping the expression data to the network data is that for some nodes there is no expression data. 
Chuang \ea \cite{Chuang2007} do not state in their paper how they handled this problem although their identified 
subnetworks contain such nodes. 
%The scoring function used to score modules in PinnacleZ during the network search 
%appears to be buggy upon addition of proteins to a module, when expression data is missing. We contacted the authors
%group about this issue but unfortunately we did not get an
%acknowledgement of the issue (personal communication). \todo{g: give
%  them one more chance, ie, write ideker directly that we will publish
%  it like this? it is a hard thing to write and they have not been
%  completely unfriendly/uncooperative}
We therefore filtered out proteins for which no expression data is
available before running PinnacleZ.
% In this study we calculate candidate subnetworks and their ranking with PinnacleZ. The calculation of the subnetwork scores 
% for a certain breast cancer study is calculated according to Equation~(\ref{activity}), where proteins for which no expression 
% data is available are filtered out.
For further issues we encountered when working with PinnacleZ see the
supplementary information.

%where $p(x, c)$ denotes the joint distribution of the elements in the activity vector $a$ and the class labels $c$, $p(x)$ and $p(c)$ denote its marginal distributions.
% Over all nodes that have a direct edge to a subnetwork $m$, this node is added that maximises the score $S_{MI}$, when added to $m$. \par
% For each of the nodes such a network is computed. The most significant networks are taken to form the feature space of a classifier. \par
% Chuang et al.\ \cite{Chuang2007} employed their method only to a protein-protein interaction network and in combination with logistic regression. Their network data stems from yeast-two-hybrid experiments.\par
% To find the optimal number of features a double-loop cross-validation was employed. They splitted the data set into five parts. Three parts were used for training, one part for validation, and one part for testing. Although determining the possible features for the classifier on an independent data set from the one used in the cross-validation they reranked the features according to the MI scores obtained for the training parts. 
% The authors made their feature extraction method available in the java package PinnacleZ and as a Cytoscape plugin.

\subsubsection*{Algorithm by Lee \ea \cite{Lee2008}}
The algorithm by Lee \ea \cite{Lee2008} uses the t-statistic to rank pathways according to their overall differential expression. 
For this it first defines sets of genes, the so called condition responsive genes (CORGs), which contain the most differentially expressed genes of a pathway. 
These genes are found by applying a greedy search. For each pathway the genes are ordered according to their t-statistics. Given the 
two sample groups let $E^1_j$ be the expression values of gene $j$ for all samples with class label $1$ and $E^0_j$ the expression values of 
gene $j$ for all samples with class label $0$, respectively. Let $k^0$ and $k^1$ denote the number of samples in 
each group; $\mu_{E^0_j}$ and $\mu_{E^1_j}$ denote the means of the two groups and $\sigma_{E^0_j}$ and $\sigma_{E^1_j}$ the standard 
deviation in the two groups.
The t-statistics between $E^1_j$ and $E^0_j$ is given by

\begin{equation}
 \label{t-test_eq}
 t(j) = \frac{\mu_{E^1_j}-\mu_{E^0_j}}{\sqrt{\sigma_{E^1_j}/k^1+\sigma_{E^0_j}/k^0}}
\end{equation}

The genes in a pathway are sorted either in ascending order, if the highest absolute $t$ value is negative, or in descending order,
if the highest absolute $t$ value is positive.
Given this order the greedy search iteratively combines genes and calculates their average expression, or \emph{pathway activity}, across the samples 
as it is given in Equation~(\ref{activity}), \ie $a_{G^\prime} = (a_{G^\prime, 1}, \dots, a_{G^\prime, k})^T$ is calculated where $G^\prime$ contains the best $m$ genes according to the ranking. 
To evaluate the combined discriminative power of the genes that have been averaged, the t-statistic of the averaged expression is 
computed as follows:
\begin{equation}
 \label{t-test_activity}
 s_{G^\prime} = 
t(G^\prime) = 
\frac{\mu_{a^1_{G^\prime}}-\mu_{a^0_{G^\prime}}}{\sqrt{\sigma_{a^1_{G^\prime}}/k^1+\sigma_{a^0_{G^\prime}}/k^0}}
\end{equation}
where $\mu_{a^0_{G^\prime}}$ and $\mu_{a^1_{G^\prime}}$ represent the means and $\sigma_{a^0_{G^\prime}}$ and $\sigma_{a^1_{G^\prime}}$ 
represent the standard deviations of the averaged activities.
If the resulting value $t(G^\prime)$ is higher than the previous value of the t-statistics 
then the search continues adding the gene to the already determined CORGs, otherwise the search stops. 
The score $s_{G^\prime}$ of the final CORGs $G'$ is then used to rank the pathway. 

As mentioned beforehand, \Lee can only be executed on predefined gene sets. Those gene sets are normally not provided in a PPI database. Thus, 
we used the KEGG and MsigDB databases to evaluate this algorithm.
In order to decrease the running time the authors executed a pathway ranking by employing the algorithm by Tian \ea 
\cite{Tian2005} and just taking the top 10\% of pathways into account prior to executing their algorithm. In our setting we ranked all of the pathways 
according to the algorithm by Lee \ea \cite{Lee2008} and considered for determining the optimized number of features in 
the final classifier all pathways in KEGG and the top 400 pathways in MsigDB\@. As feature values for classification the pathway activity, as computed according to 
Equation~(\ref{activity}) for all condition-responsive genes (CORGs), is employed. 
Here again we excluded proteins in the pathways for which no expression data is available.

\subsubsection*{Algorithm by Taylor \ea \cite{Taylor2009}}
In contrast to the algorithms by Chuang \ea \cite{Chuang2007} and Lee \ea \cite{Lee2008}, the algorithm by Taylor et 
al.\ \cite{Taylor2009} first identifies organizer proteins in the network, the so-called hubs, and then attaches a weight to 
each edge between the hubs and their direct neighbors in the network. These weights are later used to train a classifier. 

Candidate hubs are the 15\% most densely connected proteins in the network data, independent from their 
expression status. For the following calculations  proteins without expression data are excluded.
To identify hubs that significantly change their interaction behaviour between the two classes the authors introduce the hub difference and the 
average hub difference which are based on the Pearson correlation. The Pearson correlation between a hub $h$ and an interactor 
$n$ of this hub is defined as the Pearson correlation between their expression profiles across the $k$ samples
\begin{equation}
 P(h, n) = \frac{\sum^{k}_{i = 1}(e_{n, i} - \mu_{E_n})(e_{i, h} - \mu_{E_h})}{(k - 1)\sigma_{E_n}\sigma_{E_h}}\enspace.
\end{equation}

$E_n$ and $E_h$ denote the distribution of expression values across the $k$ samples and $\mu$ and $\sigma$ are their 
means and standard deviations. The hub difference is defined as the difference of the Pearson correlation $P(h, n)$ given 
the two sample classes, indicated by the superscript $0$ and $1$,%%
\begin{equation}
\label{hubdiff}
 \mathit{d}(h, n) = P^0(h, n)-P^1(h, n)\enspace .
\end{equation}

%With the help of Equation~(\ref{hubdiff}) we can now define the average hub difference. 
Let $\delta(h)$ denote the set of neighbors of a hub $h$ then the average hub difference is%%
\begin{equation}
\mathit{\overline{d}}(h) = \frac{\sum_{n\in \delta(h)} |\mathit{d}(h,n)|}{|\delta(h)|}\enspace .
\end{equation}%%

To extract only those hubs that show a significant average hub difference the value is compared to a distribution of 
the average hub difference for a permuted dataset, using a p-value cut off of $0.05$. This distribution is calculated by 1) randomly permuting the class labels and 2) recalculating 
the average hub difference and repeating these two steps 1000 times. 
The significant hubs are ranked by their average hub difference. 

As feature values in the classifier differences of the expression of the hub and each of its interactors were used. For example,  
suppose $h_1, \dots, h_p$ were found significant and suppose ${(h_1, i_{1, 1}), \dots, (h_1, i_{1, m_1}), \dots, (h_p, i_{p, 1}), \dots, (h_p, i_{p, m_p})}$ are the edges to their interactors. Then for one sample the vector $(e_{h_1} - e_{i_{1, 1}}, \dots, e_{h_1} - e_{i_{1, m_1}}, \dots, e_{h_p} - e_{i_{p, 1}}, \dots, e_{h_p} - e_{i_{p, m_p}})$ contains the feature values for the classifier.

Since the edges attached to a hub are not ranked, all those edges were included in the classifier, given that the hub 
shows a significant average hub difference. For the cross-validation procedure this means that we can not train the number of 
features but only a number of feature sets. 

\subsubsection*{Classifiers}
In our study we employed a nearest mean classifier (NMC) and logistic regression (LOG). As distance metric for the NMC
we employed the Euclidean distance. More specifically, a sample is projected on the line connecting the two class means, and the Euclidean distance 
of he projected sample to each class mean is computed. The sample is assigned to the closest class.
%of an NMC employing the Euclidean distance of the projected data points.
%The NMC is depending on the distance 
%metric with which the distance to the two class means is calculated. We implemented four versions of distances; two 
%based on the Euclidean distance, one employing the cosine angle and one projecting the data points to the straight line connecting 
%the two means and then applying the Euclidean distance. We found that there are only minor differences between 
%classification performance of NMCs employing the different distance metrics (data not shown). In our study we only present the results 
%of an NMC employing the Euclidean distance of the projected data points.

In addition to the NMC we executed all features extraction methods in combination with the LOG\@. We found that simple LOG without any 
regularization parameters cannot be trained properly since for higher numbers of features (approximately 50 features and more) 
the training step does not converge on the breast cancer data. Moreover, we found that for many features different implementations of 
LOG return different weight vectors. Thus, we used three different implementations (the R GLM, R NNET and Python SciKits implementation) and 
only accepted the classification result of the R GLM implementation when all three versions converged to the same weight vector.

\subsubsection*{Cross-validation and classification}
\begin{figure}[tbp]
\begin{center}
\includegraphics[width=9cm]{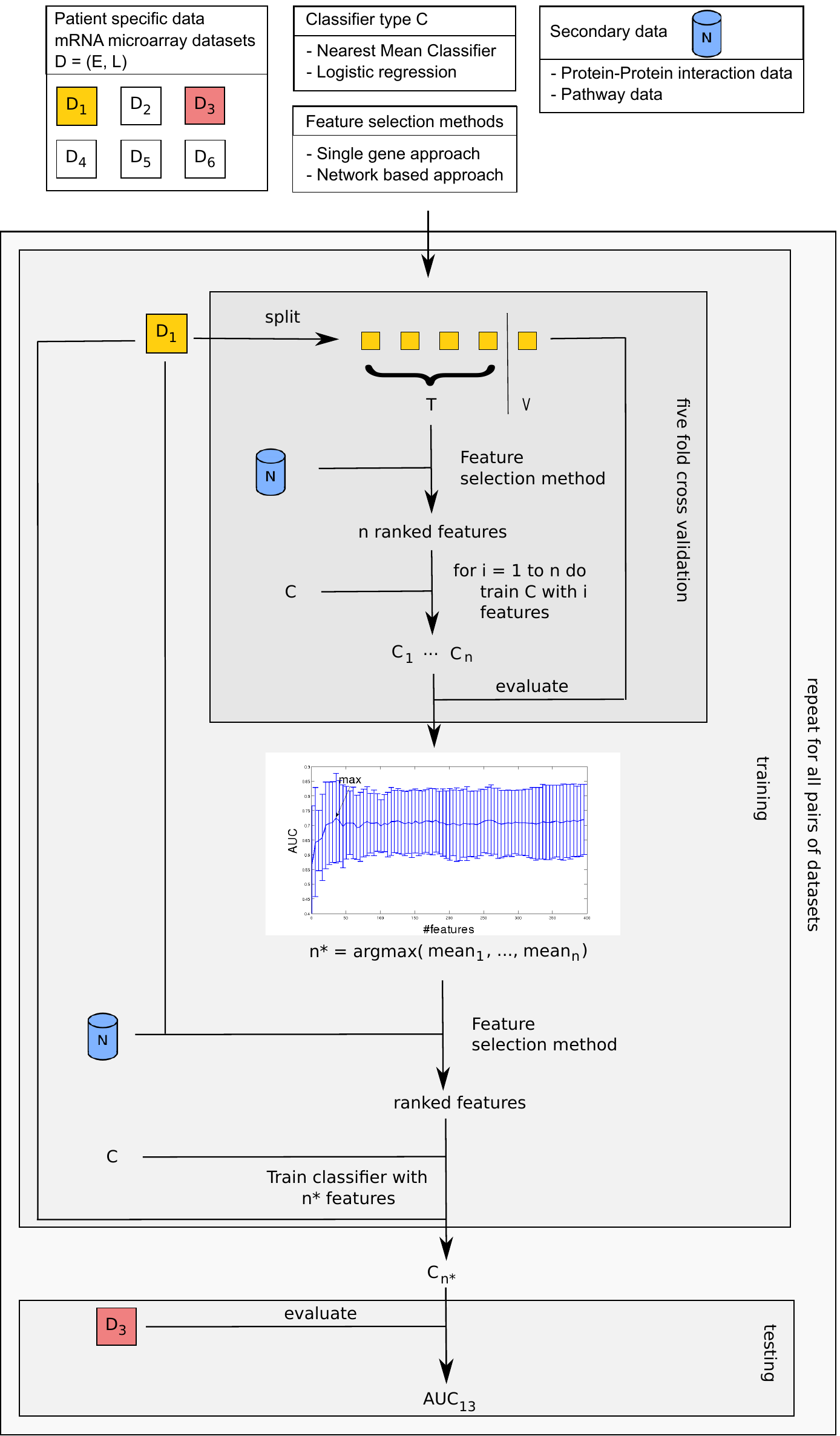}
\end{center}
\caption{\small{%%
\textbf{Overview of validation procedure.} Input: patient-specific mRNA expression data 
$D=(E, L)$ where $E$ is an expression matrix and $L$ a vector with class labels; an untrained 
classifier; a feature extraction method and an additional secondary
data source $N$ is used in the feature 
extraction method. Five fold cross-validation is used to determine 
the best performing number of features in the final classifier. The data set $D_1$ is split into five parts 
of which four parts are used as training set $T$ and one part is used as test or validation set $V$. First $T$ 
is used to extract features and to rank them. A series of classifiers is then trained, where the number of features is gradually 
increased by adding features according to the ranking. All classifiers are evaluated using $V$. 
As performance measure the area under the curve (AUC) is used.
At the end of the cross-validation each of the five splits has been used for validation once, thus we obtain five different AUC values.
We choose the number of features $n^*$ in the final classifier to be the number of features corresponding to the highest mean 
AUC value. The best $n^*$ features are determined on the whole data set $D_1$ and the final classifier with these features 
is also trained on $D_1$. This classifier is then tested on a different data set $D_3$. We evaluated all possible pairs of data sets ($\{(D_1, D_2), \dots,(D_1, D_6), (D_2, D_1), \dots, (D_6, D_5)\}$) to evaluate each feature 
extraction method in a certain setting resulting in 30 AUC values for each setting. These values were then analyzed in a box plot or a winchart. 
The features in the six classifiers ($C_{n^*}$ of each data set) were used to determine the stability of gene markers across different data sets. }}
\label{fig:cv}
\end{figure}

In the cross-validation procedure we employed, we rigorously separate the training and  
test data sets. For details, see Figure~\ref{fig:cv} and Algorithm~\ref{algo:cv}. 
The training phase consists of determining the best performing number of features and training the final classifier with this number of features. The trained classifier is then tested in the test phase. The data sets used in these two phases are completely independent, \ie the test set is never used in training the classifier.

To determine the optimized number of features in our classifiers we employed five fold cross-validation. In this cross-validation, we first
determined all required composite features (if necessary) and their ranking on four splits of the training data set. Then a series of classifiers is trained on the same four splits by gradually adding features according to the ranking. These classifiers are then evaluated on the fifth split of the data set. Since this is done in a five fold cross-validation we obtain for each of the classifiers five different AUC values. The optimized number of features extracted corresponds to the number of features yielding the highest mean performance. Once the best performing number of features is determined, the features are calculated using the whole training data set and the final classifier is trained also using the complete training data set. The classifier is then tested on the test data set. For each method we used all possible pairs of data sets as training and test set respectively. Since we have six data sets available this resulted 
in 30 AUC values for each method.

%As classifiers we employed logistic regression (LOG), which has also
%been used in Chuang \ea \cite{Chuang2007} and Lee \ea \cite{Lee2008}, and
%the nearest mean classifier (NMC).

%\obacht{figure 1 partly unclear. input $D = (D_1, \ldots,
 % D_k)$. mention algorithm 1 in figure. consistent variable names.}

%\obacht[gunnar]{Throughout the document. ``expert knowledge''. is this
%  the right word? I wouldnt use it for networks etc.}

% \obacht{algorithm 1. replace ``determine subnetworks'' by something
%   more general that applies to all methods. same: rankedNWs hints at
%   networks. rankedF better (general). same: ``networks'' (unfortunate)\\
% also: ``optimal'' classifier. unfortunate word} 

 \begin{algorithm}
\caption{Cross-validation procedure}
 \label{algo:cv}
 \begin{algorithmic}[1]
 \REQUIRE{%%
 \textbf{Datasets} $D_1 = (E_1, L_1)$, $D_2=(E_2, L_2)$ where $E_1,
 E_2$ are expression matrices and $L_1, L_2$ are vectors of outcome labels \\
\textbf{Feature extraction method} $m \in \{$SG, L, C, T$\}$\\
 \textbf{Secondary data source} $N$\\ 
 \textbf{Classifier} $C \in $ \{logistic regression, nearest mean classifier\}
 }\\
\COMMENT{Cross-validation on $D_1$}
 \STATE $(d_1, d_2, d_3, d_4, d_5) \leftarrow D_1$ \COMMENT{split samples in $D_1$ into $5$ parts $d_k = (e_k, l_k)$ where $e_k$ 
denotes the expression values of the samples in split $k$ and $l_k$ their class labels}
 \FOR{$s \leftarrow 1$ to $5$}
\STATE $\mathit{Val} \leftarrow d_s$ \COMMENT{validation set $(e_{\mathit{val}}, l_{\mathit{val}})$}
 \STATE $\mathit{Train} \leftarrow D_1 \setminus d_s$ \COMMENT{training set $e_{\mathit{train}}, l_{\mathit{train}}$}
 \STATE $\mathit{rankedF} \leftarrow m(\mathit{Train}, N)$ \COMMENT{determine features according to the published methods and rank them according to 
 the appropriate score or determine the ranking of the single genes according to the t-statistic on $\mathit{Train}$}
 \STATE $n \leftarrow $ number of $\mathit{rankedF}$
\STATE $\mathit{AUC}_s$ = empty list
 \FOR{$i \leftarrow 1$ to $n$}
 \STATE $C_{s, i} \leftarrow \mathit{trainCl}(C,
 \mathit{rankedF}[1, \dots, i],
\mathit{Train})$ \COMMENT{train the classifier with $i$ features}
 \STATE $\hat p \leftarrow C_{s, i}(e_{\mathit{Val}})$
\COMMENT{ranking of the samples in the validation set}
 \STATE $\mathit{AUC}_{s, i} \leftarrow \mathit{calcAUC}(\hat p,
 l_\mathit{Val})$ \COMMENT{calculate AUC value for $C_{s, i}$ }
 \ENDFOR
 \ENDFOR\\
\COMMENT{train CV-optimized classifier on $D_1$ and validate it on an independent data set $D_2$}
 \FOR{$j \leftarrow 1$ to n
 %$\underset{i = 1, \ldots, 5}{\min}(\mathit{length}(\mathit{AUC}_i))$
}
 \STATE $\mathit{meanAUC}[j] \leftarrow mean(\mathit{AUC}_1[j], \cdots, \mathit{AUC}_5[j])$
 \ENDFOR
 \STATE $n^* \leftarrow \argmax_j \mathit{meanAUC}[j]$ 
 \STATE $\mathit{networks} \leftarrow m(D_1, N)$
 \STATE $C_{n^*} \leftarrow \mathit{trainCl}(C, \mathit{rankedNWs}[1, \ldots, n^*], \mathit{D_1})$
 \STATE $\hat p \leftarrow C_{n^*}(E_{2})$ 
 \STATE $\mathit{AUC} \leftarrow \mathit{calcAUC}(\hat p, L_2)$
 \RETURN $\mathit{AUC}$
 \end{algorithmic}
 \end{algorithm}

\bibliography{literatur}

%\section*{Figure Legends}

%\section*{Tables}

%\begin{table}[!ht]
%\caption{
%\bf{Table title}}
%\begin{tabular}{|c|c|c|}
%table information
%\end{tabular}
%\begin{flushleft}Table caption
%\end{flushleft}
%\label{tab:label}
% \end{table}

\end{document}